\documentclass[11pt]{article}
\usepackage{soul}
\usepackage[normalem]{ulem}
\usepackage[final]{acl}
\usepackage{times}
\usepackage{latexsym}
\usepackage[T1]{fontenc}
\usepackage[utf8]{inputenc}
\usepackage{microtype}
\usepackage{inconsolata}
\usepackage{graphicx}
\usepackage{caption}
\usepackage{amsmath}
\usepackage{amssymb}
\usepackage{amsthm}
\usepackage{booktabs}
\usepackage{multirow}
\usepackage[table]{xcolor}
\definecolor{plancolor}{RGB}{255,182,193}
\definecolor{golddoccolor}{RGB}{144,238,144}
\definecolor{reasoncolor}{RGB}{173,216,230}
\definecolor{answercolor}{RGB}{255,255,224}
\usepackage{enumitem}
\usepackage{algorithm}
\usepackage{algorithmic}
\usepackage{url}

  \title{Does Faithfulness-Guided Alignment Hurt Accuracy? Unlocking Accurate and Faithful Post-Retrieval Reasoning}

\author{
\textbf{Yu Liu}\textsuperscript{1, 2}\thanks{Equal contribution.},
\textbf{Wenxiao Zhang}\textsuperscript{3}\protect\footnotemark[1],
\textbf{Diandian Guo}\textsuperscript{1, 2}\protect\footnotemark[1],
\textbf{Cong Cao}\textsuperscript{1},
\textbf{Fangfang Yuan}\textsuperscript{1}\thanks{Corresponding authors.},
\textbf{Qiang Sun}\textsuperscript{3},\\
\textbf{Yanbing Liu}\textsuperscript{1, 2},
\textbf{Jin Bum Hong}\textsuperscript{3},
\textbf{Zhiyuan Ma}\textsuperscript{4}\protect\footnotemark[2] \\
\textsuperscript{1}Institute of Information Engineering, CAS,
\textsuperscript{2}School of Cyber Security, UCAS,\\
\textsuperscript{3}The University of Western Australia,
\textsuperscript{4}Huazhong University of Science and Technology \\
\texttt{yuanfangfang@iie.ac.cn},\;
\texttt{mzyth@hust.edu.cn}
}

\begin{document}
\maketitle

\begin{abstract}
Retrieval-augmented generation (RAG) can achieve strong answer accuracy on multi-hop questions, but outcome-level rewards often leave reasoning traces weakly grounded and difficult to audit.
Under noisy retrieval, models may exhibit \emph{right-answer-wrong-reason} failures, where the final answer is correct but the supporting rationale exploits shortcuts or unsupported evidence.
We therefore ask whether faithfulness-guided alignment hurts answer accuracy in post-retrieval reasoning.
To study this question, we propose \textbf{CRAFT} (\textbf{C}alibrated \textbf{R}easoning with \textbf{A}nswer-\textbf{F}aithful \textbf{T}races), a reinforcement learning framework for the response-generation stage of retrieval-augmented multi-hop question answering.
CRAFT trains models to produce structured reasoning traces with configurable auditability, while combining deterministic rewards for format compliance, answer correctness, and citation validity with a judge-based reward for semantic faithfulness.
Experiments across model scales and benchmarks show that CRAFT unlocks task-specific reasoning capacity from 1.5B upward, improving both answer accuracy and Faithfulness; at 0.5B, performance remains sharply template-dependent.
At 7B, CRAFT improves Faithfulness over the Base model in all evaluated settings and remains competitive with strong closed-source models.
Code is available at \url{https://github.com/Ameame1/CRAFT}.
\end{abstract}

\section{Introduction}
\label{sec:intro}
\begin{figure}[t]
    \centering
    \includegraphics[width=0.46\textwidth]{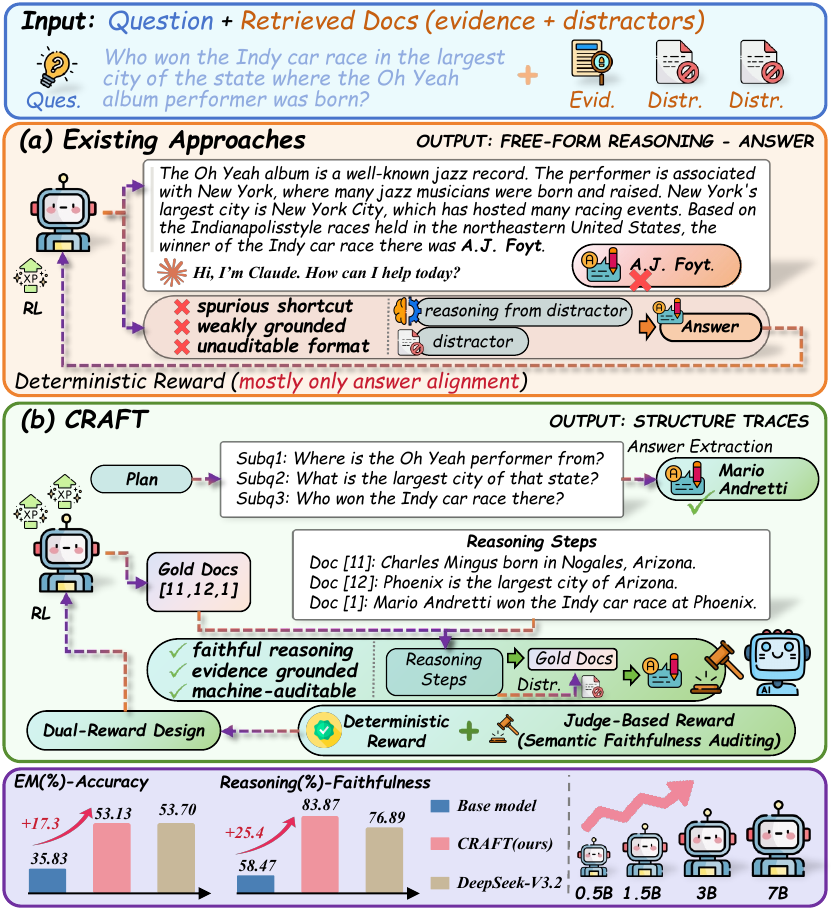}
    \caption{Comparison of existing approaches (top) and CRAFT (bottom) for multi-hop QA. Existing methods produce free-form reasoning that may exploit spurious shortcuts or lack evidence grounding. CRAFT generates structured, machine-auditable reasoning traces supervised by a dual-reward design.}
    \label{fig:motivation}
\end{figure}

Retrieval-augmented generation (RAG) grounds large language models (LLMs) in documents and is widely used for multi-hop question answering (QA), where answers require composing information across passages~\cite{lewis2020retrieval}. However, a correct answer does not imply faithful reasoning: models can exploit spurious shortcuts---citing irrelevant documents, confusing entities, or following logically inconsistent chains---yet still reach the gold answer (Figure~\ref{fig:motivation}, top). Such \emph{right-answer-wrong-reason} failures are invisible to metrics that evaluate only the final output. The problem is especially acute in multi-hop QA, which demands compositional inference over dependent steps (e.g., resolving a bridge entity and linking its attributes to downstream facts) and is further destabilized by distractors that share surface overlap with the query~\cite{trivedi2020multihop,press2023measuring}.

Prior work approaches this problem from three directions, but none directly optimizes faithful post-retrieval reasoning traces. \textbf{\textit{(1) Reasoning without faithfulness guarantees.}} Chain-of-thought and decomposition-based methods encourage step-by-step reasoning~\cite{wei2022chain,khot2023decomposed,zhou2023leasttomost,yao2023react}, but the resulting traces are not necessarily grounded in retrieved evidence; under noisy retrieval, models may bypass relevant documents and produce plausible-looking but unfaithful rationales~\cite{lanham2023measuring,turpin2023language,paul2024making}. \textbf{\textit{(2) Grounding as post-hoc verification.}} Citation, evaluation, and verification mechanisms introduce evidence signals such as document-level citations or consistency checks~\cite{gao2023enabling,es2024ragas,wallat2025correctness,tutek2025measuring,arakelyan2025flare,sui2025fidelis}, yet they generally evaluate completed outputs or operate at the answer level, leaving intermediate reasoning difficult to audit at scale. \textbf{\textit{(3) RL without fixed-evidence trace supervision.}} Reinforcement learning methods optimize retrieval, reasoning, or generation using answer- and relevance-based rewards~\cite{asai2024selfrag,li2025r3rag,chen2025mmoa}, but primarily supervise retrieval/search trajectories or final outcomes rather than the semantic faithfulness of structured answer-generation traces over a fixed evidence set.

To address these limitations, we propose \textbf{CRAFT} (\textbf{C}alibrated \textbf{R}easoning with \textbf{A}nswer-\textbf{F}aithful \textbf{T}races), a reinforcement learning framework based on Group Relative Policy Optimization (GRPO)~\cite{shao2024deepseekmath} for the response generation stage of retrieval-augmented multi-hop QA. CRAFT defines a family of structured output templates that decompose generation into auditable stages (planning, evidence selection, step-by-step reasoning, and answer extraction), forming trace variants with configurable levels of auditability (Figure~\ref{fig:motivation}, bottom). Within each variant, these stages form a \emph{chain of faithfulness} where each component constrains the next, so that unfaithful reasoning is more likely to violate at least one verifiable link. Training combines two complementary forms of supervision: deterministic rewards enforce format validity, citation-set correctness, and answer matching, while a judge-based reward from a high-capacity LLM audits semantic faithfulness by verifying cross-stage consistency (plan$\rightarrow$reason, citation$\rightarrow$reason, reason$\rightarrow$answer) and grounding individual claims against the retrieved evidence. This extends process-level supervision beyond rule-based objectives to open-ended semantic properties (e.g., reasoning consistency and claim verifiability). The judge operates on binary decomposed criteria to limit bias propagation, and its reliability is assessed against human annotations (\S\ref{sec:human_validation}). Together, these designs enable the model to learn not just \emph{what} to answer but \emph{how} to reason faithfully. Our contributions are as follows:

\begin{itemize}[leftmargin=*]
    \item We propose \textbf{CRAFT}, an RL framework that unlocks task-specific multi-hop reasoning through controllable, auditable trace variants (\textsc{CRAFT}$_{\text{v1}}$--\textsc{CRAFT}$_{\text{v5}}$).
    \item We introduce a \textbf{dual-reward design} that combines \textit{deterministic rewards} for verifiable constraints with \textit{judge-based rewards} for reasoning faithfulness, extending process-level supervision from rule-based objectives to open-ended semantic properties.
    \item Experiments across model scales and benchmarks show that faithfulness-guided alignment usually yields modest answer-performance gains rather than hurting accuracy. Full CRAFT improves \textbf{Faithfulness over Base in all 12 evaluated 7B cells} and remains competitive with strong closed-source models.
\end{itemize}

\section{Related Work}
\label{sec:related}

\subsection{Faithful RAG Reasoning}

CoT prompting~\cite{wei2022chain} and decomposition-based reasoning can improve multi-step inference, and rationale generation has also been studied in multimodal settings~\cite{qin2024genhmd}. However, generated rationales may not faithfully reflect the model's decision process~\cite{lanham2023measuring,turpin2023language,paul2024making,zhang2025reliability}; recent work probes sustained multi-turn reasoning behavior~\cite{zhang2025cogmem} and separates step relevance, evidence attribution, and logical correctness in fine-grained verification~\cite{jacovi2024weakest}. RAG~\cite{lewis2020retrieval} grounds generation in documents, with multi-hop extensions such as IRCoT~\cite{trivedi2023interleaving}, HopRAG~\cite{liu2025hoprag}, and OPERA~\cite{liu2025opera}. Faithfulness-oriented evaluation and verification methods, including RAGAs~\cite{es2024ragas} and Generate but Verify~\cite{filice2025generate}, assess whether generated answers are supported by retrieved context, while studies on citations, attribution, factuality, and grounding-oriented rewards show that citation support or factual correctness alone may not ensure context-faithful reasoning~\cite{gao2023enabling,wallat2025correctness,bi2025factuality,wang2025reasoningretrieval,wei2025instructrag,li2025ragddr}. These lines motivate auditable reasoning in RAG, but mostly evaluate completed outputs or answer-level grounding; we focus on training structured post-retrieval traces with both structural and semantic faithfulness checks.

\begin{figure*}[t]
    \centering
    \includegraphics[width=\textwidth]{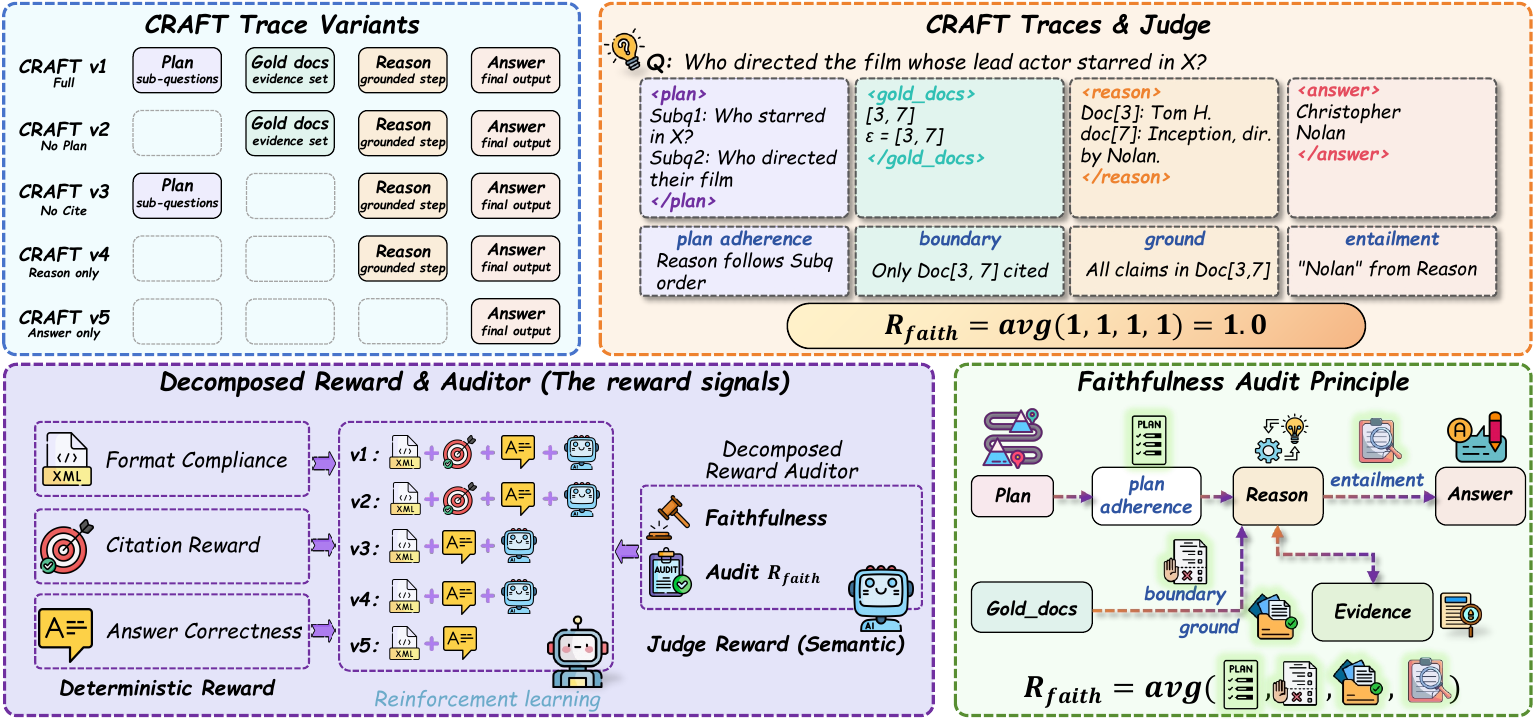}\
    \caption{\textbf{Overview of CRAFT.} The framework comprises three components: CRAFT Trace Variants (top-left) defines structured trace templates (\textsc{CRAFT}$_{\text{v1}}$--\textsc{CRAFT}$_{\text{v5}}$); Decomposed Reward Auditor (bottom-left) provides four reward signals ($R_{\mathrm{fmt}}$, $R_{\mathrm{gold}}$, $R_{\mathrm{faith}}$, $R_{\mathrm{ans}}$); Faithfulness Audit Principle (bottom-right) enables machine-checkable reasoning verification. Top-right: a concrete \textsc{CRAFT}$_{\text{v1}}$ trace with all four judge checks passing ($R_{\mathrm{faith}}=1.0$).}
    \label{fig:method_overview}
\end{figure*}

\subsection{Reinforcement Learning for Reasoning}

Preference alignment includes RLHF with PPO~\cite{ouyang2022training,schulman2017proximal} and direct preference methods such as DPO~\cite{rafailov2023direct} and RRHF~\cite{yuan2023rrhf}; GRPO~\cite{shao2024deepseekmath} removes the critic via group-wise normalization and has been used for reasoning models~\cite{deepseekr1}. Process rewards can improve exploration efficiency~\cite{setlur2025rewarding}, but often rely on domain-specific verifier data. In RAG, multi-agent RL~\cite{chen2025mmoa} and R3-RAG's answer-correctness and document-relevance rewards optimize interleaved reasoning and retrieval~\cite{li2025r3rag}. LLM-as-a-Judge provides scalable reward or evaluation signals~\cite{zheng2023judging}, although it can introduce biases~\cite{chen2024humans,gu2024survey}; common mitigations include binary criteria, rubric-based scoring, and human validation~\cite{adlakha2024evaluating,bavaresco2024llms}. However, existing RL and judge-based approaches rarely target fixed-evidence answer generation with structured trace-level faithfulness checks. We use these ideas to combine deterministic rewards with decomposed judge checks over post-retrieval reasoning traces.

\subsection{Multi-hop Question Answering}
Multi-hop QA benchmarks~\cite{yang2018hotpotqa,ho2020constructing,trivedi2022musique} require composing information across documents, yet models exhibit a compositionality gap~\cite{press2023measuring} and exploit shortcuts~\cite{trivedi2020multihop}. Decomposition-based~\cite{khot2023decomposed,zhou2023leasttomost} and agentic~\cite{yao2023react} methods improve performance, while structured multi-step agents have also been explored in related alignment tasks~\cite{nan2026ea}. Fine-grained evaluation shows final-answer scores can conceal faulty intermediate reasoning~\cite{liu2025beyondanswer}. Recent work further examines implicit reasoning~\cite{yao2025implicit}, back attention~\cite{yu2025back}, and difficulty-controlled evaluation~\cite{lee2025grade}; these studies motivate evaluation beyond answers, especially for evidence-grounded reasoning traces.

\section{Method}
\label{sec:method}

\subsection{Problem Formulation}
\label{sec:problem}

We study \textbf{the response generation stage of RAG-based LLMs} for multi-hop QA. Each instance contains a question $q$ and retrieved document set $D=\{d_1,\ldots,d_K\}$ that \emph{mixes evidence with distractors}. Answering $q$ requires reasoning over supporting documents indexed by $\mathcal{S}^*\subseteq\{1,\ldots,K\}$, while distractors introduce spurious shortcuts. Our goal is to train models that generate both correct answers $a$ and auditable reasoning traces: XML-structured outputs (\texttt{<plan>}, \texttt{<gold\_docs>}, \texttt{<reason>}, \texttt{<answer>}) that expose derivation. Formally, given $(q,D)$ and a gold answer $a^*$, the model outputs a trace $y$ with components $\pi$ (plan), $\mathcal{E}$ (citation set), $\rho$ (reasoning), and $a$ (answer), satisfying: (i) format parsability $\mathcal{F}(y)=1$, (ii) answer correctness $a=a^*$, and (iii) trace faithfulness $\mathcal{C}(\pi,\mathcal{E},\rho,a)=1$, where $\mathcal{C}$ enforces internal consistency and evidence grounding.

\subsection{Overview}
\label{sec:overview}

As illustrated in Figure~\ref{fig:method_overview}, CRAFT consists of three tightly integrated components:
(1) \textbf{CRAFT Trace Variants} that define structured output templates (top),
(2) a \textbf{Decomposed Reward Auditor} that provides multi-signal supervision (center), and
(3) a \textbf{Faithfulness Audit Principle} that enables machine-checkable reasoning verification (right).

\noindent\textbf{Trace Templates.}
We define a family of XML-based output templates spanning an auditability spectrum.
The full template (\textsc{CRAFT}$_{\textbf{v1}}$) decomposes model output into four fields forming an auditable chain:
\texttt{<plan>} declares the reasoning blueprint (sub-question decomposition);
\texttt{<gold\_docs>} commits to an evidence boundary \emph{before} reasoning begins;
\texttt{<reason>} must cite \emph{only within} this boundary while addressing the plan;
\texttt{<answer>} must be logically entailed by the reasoning.
Intermediate variants selectively ablate components:
\textsc{CRAFT}$_{\textbf{v2}}$ removes \texttt{<plan>},
\textsc{CRAFT}$_{\textbf{v3}}$ removes \texttt{<gold\_docs>},
\textsc{CRAFT}$_{\textbf{v4}}$ retains only \texttt{<reason>}+\texttt{<answer>},
and \textsc{CRAFT}$_{\textbf{v5}}$ (Answer-Only) serves as a minimal baseline.
This design creates a \textbf{chain of faithfulness}: $\pi \rightarrow \mathcal{E} \rightarrow \rho \rightarrow a$, where unsupported or inconsistent content can be exposed through explicit links. The template also determines which rewards apply: variants containing \texttt{<gold\_docs>} activate $R_{\mathrm{gold}}$, while variants with reasoning fields activate $R_{\mathrm{faith}}$.

\subsection{CRAFT Training}
\label{sec:craft}

\noindent\textbf{Principle.}
\textit{We optimize the policy $\pi_\theta$ to generate traces that are not only correct but structurally valid and reason-faithful, effectively supervising the ``chain of faithfulness''.}
For each input $x=(q,D,v)$, GRPO samples a group of $G$ traces $\{y_i\}_{i=1}^{G}\sim \pi_\theta(\cdot\mid x)$ and normalizes rewards within the group:
\begin{equation}
\hat{R}_i = \frac{R(y_i)-\mu}{\sigma+\epsilon},
\quad
\mu=\frac{1}{G}\sum_{j=1}^{G}R(y_j),
\end{equation}
where $\sigma$ is the standard deviation of $\{R(y_j)\}_{j=1}^{G}$ and $\epsilon$ is a small constant for numerical stability.
The algorithm then performs the critic-free, token-level clipped update used by our GRPO implementation. Let
$r_{i,t}(\theta)=\pi_\theta(y_{i,t}\mid x,y_{i,<t})/\pi_{\theta_{\mathrm{old}}}(y_{i,t}\mid x,y_{i,<t})$ and let $D_{i,t}$ denote the per-token KL estimator against the reference policy. The maximized objective is
\begin{align}
\bar r_{i,t}(\theta)
&=\operatorname{clip}(r_{i,t}(\theta),1-\varepsilon,1+\varepsilon),\\
\ell^{\mathrm{clip}}_{i,t}(\theta)
&=\min\!\left\{r_{i,t}(\theta)\hat R_i,\,
\bar r_{i,t}(\theta)\hat R_i\right\},\\
\mathcal{L}_i(\theta)
&=\frac{1}{T_i}\sum_{t=1}^{T_i}\!\left(
\ell^{\mathrm{clip}}_{i,t}(\theta)-\beta D_{i,t}\right),\\
\mathcal{J}(\theta)
&=\mathbb{E}_{x}\!\left[\frac{1}{G}\sum_{i=1}^{G}
\mathcal{L}_i(\theta)\right].
\end{align}
We use $\varepsilon=0.2$ and $\beta=0.04$.
The key design choice is the total reward $R(y)$, which we decompose into four components:
\begin{equation}
\begin{aligned}
R(y) &=
\sum_{c \in \mathcal{C}_v} w_c\, R_c(y), \\
\mathcal{C}_v &\subseteq \{\mathrm{fmt}, \mathrm{gold}, \mathrm{faith}, \mathrm{ans}\},
\end{aligned}
\end{equation}
where weights $w_{(\cdot)}$ control the trade-off between strictness and accuracy. The implementation sums the weighted reward terms, so $R(y)\in[0,\sum_{c\in\mathcal{C}_v}w_c]$ when each component lies in $[0,1]$.

\subsubsection{Deterministic Rewards}
Rewards of this type can be computed directly from the trace structure and gold labels without requiring an LLM judge.

\noindent\textbf{Format Compliance ($R_{\mathrm{fmt}}$).}
\textit{We enforce strict adherence to the XML schema to ensure parsability.}
\begin{equation}
R_{\mathrm{fmt}}(y) = \mathbb{I}[y \in \mathcal{V}_{v}],
\end{equation}
where $\mathcal{V}_{v}$ is the set of strings containing all required tags for variant $v$ in correct order. The deterministic answer and citation rewards are gated by this check and return $0$ when $R_{\mathrm{fmt}}(y)=0$; the judge reward is computed independently from the submitted trace.

\noindent\textbf{Citation Reward ($R_{\mathrm{gold}}$).}
\textit{We encourage document-level citations that cover the supporting documents while avoiding distractors.}
When \texttt{<gold\_docs>} is present, we parse the predicted citation set $\mathcal{E}$ and compare it with the gold supports $\mathcal{S}^*$:
\begin{equation}
R_{\mathrm{gold}}(y) =
\begin{cases}
1, & \mathcal{E}=\mathcal{S}^*,\\
0.5, & \mathcal{E}\cap\mathcal{S}^*\neq\varnothing\ \text{and}\ \mathcal{E}\neq\mathcal{S}^*,\\
0, & \mathcal{E}\cap\mathcal{S}^*=\varnothing.
\end{cases}
\end{equation}

\noindent\textbf{Answer Correctness ($R_{\mathrm{ans}}$).}
\textit{We target final task accuracy with normalized exact-match supervision.}
\begin{equation}
\begin{aligned}
z(y)&=\textsc{Norm}\!\left(\textsc{ExtractAnswer}(y)\right),\\
\widetilde A^*&=\left\{\textsc{Norm}(a):a\in A^*\right\},\\
R_{\mathrm{ans}}(y,A^*)&=\mathbb{I}\!\left[z(y)\in\widetilde A^*\right].
\end{aligned}
\end{equation}
where $\textsc{ExtractAnswer}(y)$ parses the \texttt{<answer>} content, $A^*$ is the set of acceptable references, and $\textsc{Norm}$ lowercases text and removes punctuation, articles, underscores, and redundant whitespace.

\begin{algorithm}[t]
\captionsetup{font=footnotesize}
\footnotesize
\caption{CRAFT Faithfulness Judge (\textsc{CRAFT}$_{\text{v1}}$)}
\label{alg:judge_v1}
\begin{algorithmic}[1]
\REQUIRE Query $q$, Retrieved documents $D = \{d_1, \ldots, d_K\}$
\REQUIRE Model trace $\mathcal{T}$ with four XML fields:
\STATE $\bullet$\texttt{<plan>}: Decomposed sub-questions
\STATE $\bullet$\texttt{<gold\_docs>}: Declared evidence indices, e.g., $[2,5]$
\STATE $\bullet$\texttt{<reason>}: Step-by-step reasoning with doc citations
\STATE $\bullet$\texttt{<answer>}: Final answer to $q$
\ENSURE \mbox{Binary audit scores and one overall verdict}
 \colorbox{gray!20}{\textbf{Faithfulness Judge Principles (A, B, C, D)}}
\STATE \textbf{------- A: Plan $\rightarrow$ Reason -------}
\STATE Check: Does \texttt{reason} follow the sub-questions in \texttt{plan}?
\STATE $c_{\pi} \leftarrow \mathbf{1}[\text{reasoning addresses plan's intent in order}]$
\STATE \textbf{------- B: Gold\_docs $\rightarrow$ Reason -------}
\STATE $\mathcal{E} \leftarrow$ doc indices declared in \texttt{gold\_docs}
\STATE $\mathcal{R} \leftarrow$ doc indices actually cited in \texttt{reason}
\STATE Check: Are all citations within the declared boundary?
\STATE $c_{\mathcal{E}} \leftarrow \mathbf{1}[\mathcal{R} \subseteq \mathcal{E} \land \mathcal{R} \neq \emptyset]$
\STATE \textbf{------- C: Reason $\rightarrow$ Answer -------}
\STATE Check: Is \texttt{answer} a logical conclusion of \texttt{reason}?
\STATE $c_{a} \leftarrow \mathbf{1}[\text{answer supported by reasoning chain}]$
\STATE \textbf{------- D: Evidence $\rightarrow$ Reason (Grounding) -------}
\STATE Check: Are claims in \texttt{reason} supported by cited docs?
\STATE $c_{g} \leftarrow 1$
\FOR{each claim $k_i$ citing document $d_j \in \mathcal{E}$}
    \IF{$D[d_j]$ does not support $k_i$}
        \STATE $c_{g} \leftarrow 0$ \COMMENT{Hallucination detected}
    \ENDIF
\ENDFOR
\STATE $F_{\mathrm{eval}} \leftarrow o(\mathcal{T},D)\in\{0,1\}$ \COMMENT{Judge's overall verdict}
\RETURN training reward $R_{\mathrm{faith}}=\frac{1}{4}(c_{\pi}+c_{\mathcal{E}}+c_a+c_g)$ and evaluation verdict $F_{\mathrm{eval}}$
\end{algorithmic}
\end{algorithm}

\subsubsection{Judge Reward}
Rewards of this type rely on an LLM judge to assess semantic consistency and evidence grounding, supervising the reasoning \emph{process} rather than only the outcome.

\noindent\textbf{Faithfulness Audit ($R_{\mathrm{faith}}$).}
\textit{We penalize hallucinated reasoning by verifying internal consistency and citation grounding} (Algorithm~\ref{alg:judge_v1}).
Let $\textsc{Parse}(y)\mapsto (\pi,\mathcal{E},\rho,a)$ denote extracting the trace fields from $y$. We employ a high-capacity LLM judge to compute:
\begin{equation}
R_{\mathrm{faith}}(y) = \text{Audit}(\pi, \mathcal{E}, \rho, a; D, v),
\end{equation}
by averaging multiple machine-checkable criteria:
\begin{equation}
\text{Audit}(\cdot)=\frac{1}{m_v}\sum_{k=1}^{m_v} r_k(y),
\quad r_k \in \mathcal{R}_v,
\end{equation}
where $\mathcal{R}_v=\{r_{\pi\rightarrow\rho},\, r_{\mathcal{E}\rightarrow\rho},\, r_{\rho\rightarrow a},\, r_{\mathrm{ground}}\}$ and $m_v$ depends on the trace variant (e.g., plan/citation checks apply only when the corresponding fields exist). Concretely, the judge checks:
(i) \textbf{plan$\rightarrow$reason} ($r_{\pi\rightarrow\rho}$),
(ii) \textbf{citation-set$\rightarrow$reason} ($r_{\mathcal{E}\rightarrow\rho}$),
(iii) \textbf{reason$\rightarrow$answer} ($r_{\rho\rightarrow a}$), and
(iv) \textbf{grounding} ($r_{\mathrm{ground}}$), i.e., key claims in $\rho$ are supported by the cited document text in $D$.
Each check yields a binary score, and the judge also emits a binary overall-consistency verdict $o(y)$ informed by these criteria. During training, the component average supplies a dense reward; during evaluation, we use the final verdict:
\begin{align}
R_{\mathrm{faith}}(y)&=\frac{1}{m_v}\sum_{k=1}^{m_v}r_k(y)\in[0,1],\\
F_{\mathrm{eval}}(y)&=o(y)\in\{0,1\},\\
\mathrm{Faith}&=\frac{1}{|\mathcal{J}|}\sum_{y\in\mathcal{J}}F_{\mathrm{eval}}(y).
\end{align}
where $\mathcal{J}$ contains all judged evaluation traces. Thus, training uses graded component-level feedback, while main-table Faithfulness records one overall-consistency decision per trace. The judge model (Qwen3-30B-A3B) is distinct from the trained Qwen2.5 policies, and its reliability is assessed against human annotations (\S\ref{sec:human_validation}).

\begin{table*}[t]
    \centering
    \scriptsize
    \setlength{\tabcolsep}{3pt}
    \begin{tabular*}{\textwidth}{@{\extracolsep{\fill}}lcccccccccc@{}}
        \toprule
        \textbf{Method} & \textbf{Type} & \multicolumn{3}{c}{\textbf{MuSiQue}} & \multicolumn{3}{c}{\textbf{HotpotQA}} & \multicolumn{3}{c}{\textbf{2WikiMHQA}} \\
         &  & EM & F1 & Faith. & EM & F1 & Faith. & EM & F1 & Faith. \\
        \midrule
        \rowcolor{gray!20}\multicolumn{11}{l}{\textit{Open-source}} \\
        \quad Qwen2.5-0.5B & Pre-Trained & 0.25 & 1.32 & 0.19 & 1.76 & 5.58 & 1.69 & 6.02 & 9.95 & 1.64 \\
        \quad Qwen2.5-1.5B & Pre-Trained & 0.91 & 3.63 & 5.13 & 9.96 & 17.28 & 14.63 & 10.10 & 15.95 & 9.22 \\
        \quad Qwen2.5-3B & Pre-Trained & 12.03 & 19.62 & 24.28 & 32.61 & 44.34 & 56.00 & 30.62 & 38.65 & 44.59 \\
        \quad Qwen2.5-7B & Pre-Trained & 35.83 & 47.99 & 58.47 & 56.47 & 71.58 & 83.30 & 57.87 & 66.56 & 76.92 \\
        \quad Qwen3-4B & Pre-Trained & 24.11 & 32.40 & 52.68 & 51.30 & 65.58 & 87.81 & 54.24 & 64.33 & 86.20 \\
        \quad Qwen3-30B-A3B & Pre-Trained & 40.12 & 48.95 & 68.80 & 59.57 & 74.24 & 92.89 & 55.73 & 66.19 & 90.78 \\
        \midrule
        \rowcolor{gray!20}\multicolumn{11}{l}{\textit{Closed-source}} \\
        \quad GPT-5-mini & API & 53.59 & \uwave{68.42} & \uwave{83.42} & 61.00 & \uwave{80.22} & \uwave{96.43} & \uwave{70.47} & \uwave{81.19} & 95.98 \\
        \quad DeepSeek-V3.2 & API & \uwave{53.70} & 67.81 & 76.89 & 63.56 & 79.51 & 94.31 & 69.20 & 77.74 & \uwave{96.80} \\
        \quad Gemini-2.5-Flash & API & 52.37 & 63.62 & 77.55 & \uwave{63.60} & 78.80 & 93.41 & 69.76 & 79.35 & 94.42 \\
        \midrule
        \rowcolor{gray!20}\multicolumn{11}{l}{\textit{SFT}} \\
        \quad Qwen2.5-0.5B & Trained & 0.00 & 0.00 & 0.10 & 0.82 & 2.92 & 0.70 & 2.81 & 4.62 & 0.68 \\
        \quad Qwen2.5-1.5B & Trained & 7.44 & 14.69 & 10.14 & 36.51 & 47.95 & 33.07 & 34.49 & 42.81 & 33.03 \\
        \quad Qwen2.5-3B & Trained & 18.52 & 25.28 & 26.36 & 47.87 & 62.12 & 69.06 & 49.37 & 58.83 & 66.65 \\
        \quad Qwen2.5-7B & Trained & 31.01 & 41.78 & 51.22 & 57.81 & 71.30 & 83.37 & 62.69 & 71.50 & 79.00 \\
        \midrule
        \rowcolor{gray!20}\multicolumn{11}{l}{\textit{Ours (RL)}} \\
        \quad CRAFT$_{\text{0.5B}}$ & Trained & 0.00 & 0.79 & 1.11 & 0.44 & 2.25 & 0.88 & 1.35 & 3.71 & 1.73 \\
        \quad CRAFT$_{\text{1.5B}}$ & Trained & 26.87 & 36.00 & 28.35 & 46.35 & 59.47 & 60.27 & 43.57 & 51.19 & 48.46 \\
        \quad CRAFT$_{\text{3B}}$ & Trained & 41.44 & 49.96 & 58.69 & 56.19 & 69.92 & 78.52 & 56.75 & 63.45 & 62.11 \\
        \quad CRAFT$_{\text{7B}}$ & Trained & \underline{53.13} & \underline{62.97} & \underline{83.87} & \underline{64.06} & \underline{78.86} & \underline{96.31} & \underline{78.23} & \underline{84.18} & \underline{96.79} \\
        \bottomrule
    \end{tabular*}
        \caption{Main results (\textsc{CRAFT}$_{\text{v1}}$) on multi-hop QA benchmarks. All metrics are percentages. \uwave{Wavy underline}: best API result; \underline{straight underline}: best non-API result.}
    \label{tab:main_results_v1}
\end{table*}

\begin{table*}[t]
    \centering
    \scriptsize
    \setlength{\tabcolsep}{2.2pt}
    \begin{tabular*}{\textwidth}{@{\extracolsep{\fill}}lccccccccc@{}}
        \toprule
        \textbf{Version} & \multicolumn{3}{c}{\textbf{MuSiQue}} & \multicolumn{3}{c}{\textbf{HotpotQA}} & \multicolumn{3}{c}{\textbf{2WikiMHQA}} \\
         & EM & F1 & Faith. & EM & F1 & Faith. & EM & F1 & Faith. \\
        \midrule
        \rowcolor{gray!20}\multicolumn{10}{l}{\textit{Qwen2.5-7B$_{\text{Base}}$}} \\
        CRAFT$_{\text{v1}}$ & 35.83 & 47.99 & 58.47 & 56.47 & 71.58 & 83.30 & 57.87 & 66.56 & 76.92 \\
        CRAFT$_{\text{v2}}$ & 34.24 & 44.77 & 66.15 & 54.56 & 69.62 & 87.80 & 55.60 & 65.63 & 77.42 \\
        CRAFT$_{\text{v3}}$ & 39.11 & 49.50 & 68.47 & 58.38 & 71.91 & 90.53 & 61.07 & 69.85 & 84.52 \\
        CRAFT$_{\text{v4}}$ & 38.41 & 49.14 & 76.53 & 60.12 & 74.40 & 93.72 & 60.80 & 70.37 & 86.65 \\
        CRAFT$_{\text{v5}}$ & 31.43 & 41.91 & - & 53.38 & 67.97 & - & 51.01 & 59.21 & - \\
        \midrule
        \rowcolor{gray!20}\multicolumn{10}{l}{\textit{Qwen2.5-7B$_{\text{SFT}}$}} \\
        CRAFT$_{\text{v1}}$ & 31.01{\tiny\color{red}(-4.82)} & 41.78{\tiny\color{red}(-6.21)} & 51.22{\tiny\color{red}(-7.25)} & 57.81{\tiny\color{blue}(+1.34)} & 71.30{\tiny\color{red}(-0.28)} & 83.37{\tiny\color{blue}(+0.07)} & 62.69{\tiny\color{blue}(+4.82)} & 71.50{\tiny\color{blue}(+4.94)} & 79.00{\tiny\color{blue}(+2.08)} \\
        CRAFT$_{\text{v2}}$ & 28.38{\tiny\color{red}(-5.86)} & 40.32{\tiny\color{red}(-4.45)} & 56.00{\tiny\color{red}(-10.15)} & 55.51{\tiny\color{blue}(+0.95)} & 70.42{\tiny\color{blue}(+0.80)} & 88.14{\tiny\color{blue}(+0.34)} & 59.67{\tiny\color{blue}(+4.07)} & 67.86{\tiny\color{blue}(+2.23)} & 76.31{\tiny\color{red}(-1.11)} \\
        CRAFT$_{\text{v3}}$ & 34.55{\tiny\color{red}(-4.56)} & 46.19{\tiny\color{red}(-3.32)} & 60.29{\tiny\color{red}(-8.18)} & 58.59{\tiny\color{blue}(+0.21)} & 73.20{\tiny\color{blue}(+1.29)} & 88.45{\tiny\color{red}(-2.08)} & 66.06{\tiny\color{blue}(+4.99)} & 74.76{\tiny\color{blue}(+4.91)} & 85.71{\tiny\color{blue}(+1.19)} \\
        CRAFT$_{\text{v4}}$ & 30.11{\tiny\color{red}(-8.30)} & 41.53{\tiny\color{red}(-7.61)} & 65.80{\tiny\color{red}(-10.73)} & 57.54{\tiny\color{red}(-2.58)} & 71.84{\tiny\color{red}(-2.56)} & 89.40{\tiny\color{red}(-4.32)} & 63.92{\tiny\color{blue}(+3.12)} & 72.97{\tiny\color{blue}(+2.60)} & 82.37{\tiny\color{red}(-4.28)} \\
        CRAFT$_{\text{v5}}$ & 25.60{\tiny\color{red}(-5.83)} & 36.27{\tiny\color{red}(-5.64)} & - & 54.30{\tiny\color{blue}(+0.92)} & 68.26{\tiny\color{blue}(+0.29)} & - & 55.47{\tiny\color{blue}(+4.46)} & 63.41{\tiny\color{blue}(+4.20)} & - \\
        \midrule
        \rowcolor{gray!20}\multicolumn{10}{l}{\textit{CRAFT$_{\text{7B}}$ w/o judge reward}} \\
        CRAFT$_{\text{v1}}$ & 51.53{\tiny\color{blue}(+15.70)} & 60.88{\tiny\color{blue}(+12.89)} & 81.96{\tiny\color{blue}(+23.49)} & 62.49{\tiny\color{blue}(+6.02)} & 76.86{\tiny\color{blue}(+5.28)} & 94.81{\tiny\color{blue}(+11.51)} & 76.47{\tiny\color{blue}(+18.60)} & 82.18{\tiny\color{blue}(+15.63)} & 94.67{\tiny\color{blue}(+17.75)} \\
        CRAFT$_{\text{v2}}$ & 49.72{\tiny\color{blue}(+15.48)} & 60.65{\tiny\color{blue}(+15.88)} & 80.21{\tiny\color{blue}(+14.06)} & 63.68{\tiny\color{blue}(+9.12)} & 78.57{\tiny\color{blue}(+8.95)} & 94.66{\tiny\color{blue}(+6.86)} & 73.33{\tiny\color{blue}(+17.73)} & 80.73{\tiny\color{blue}(+15.10)} & 92.62{\tiny\color{blue}(+15.20)} \\
        CRAFT$_{\text{v3}}$ & 50.21{\tiny\color{blue}(+11.10)} & 60.46{\tiny\color{blue}(+10.96)} & 82.57{\tiny\color{blue}(+14.10)} & 62.07{\tiny\color{blue}(+3.69)} & 76.64{\tiny\color{blue}(+4.73)} & 94.27{\tiny\color{blue}(+3.74)} & 74.60{\tiny\color{blue}(+13.53)} & 81.73{\tiny\color{blue}(+11.88)} & 93.66{\tiny\color{blue}(+9.14)} \\
        CRAFT$_{\text{v4}}$ & 46.24{\tiny\color{blue}(+7.83)} & 58.42{\tiny\color{blue}(+9.28)} & 81.68{\tiny\color{blue}(+5.15)} & 60.92{\tiny\color{blue}(+0.80)} & 75.62{\tiny\color{blue}(+1.22)} & 92.96{\tiny\color{red}(-0.76)} & 73.59{\tiny\color{blue}(+12.79)} & 80.21{\tiny\color{blue}(+9.84)} & 91.29{\tiny\color{blue}(+4.64)} \\
        CRAFT$_{\text{v5}}$ & 49.95{\tiny\color{blue}(+18.52)} & 60.51{\tiny\color{blue}(+18.60)} & - & 59.80{\tiny\color{blue}(+6.42)} & 74.79{\tiny\color{blue}(+6.82)} & - & 60.34{\tiny\color{blue}(+9.33)} & 67.72{\tiny\color{blue}(+8.51)} & - \\
        \midrule
        \rowcolor{gray!20}\multicolumn{10}{l}{\textit{CRAFT$_{\text{7B}}$ (full)}} \\
        CRAFT$_{\text{v1}}$ & 53.13{\tiny\color{blue}(+17.30)} & 62.97{\tiny\color{blue}(+14.99)} & 83.87{\tiny\color{blue}(+25.40)} & 64.06{\tiny\color{blue}(+7.59)} & 78.86{\tiny\color{blue}(+7.28)} & 96.31{\tiny\color{blue}(+13.01)} & 78.23{\tiny\color{blue}(+20.36)} & 84.18{\tiny\color{blue}(+17.63)} & 96.79{\tiny\color{blue}(+19.87)} \\
        CRAFT$_{\text{v2}}$ & 51.31{\tiny\color{blue}(+17.07)} & 63.06{\tiny\color{blue}(+18.29)} & 82.49{\tiny\color{blue}(+16.34)} & 65.28{\tiny\color{blue}(+10.72)} & 80.10{\tiny\color{blue}(+10.48)} & 96.38{\tiny\color{blue}(+8.58)} & 75.26{\tiny\color{blue}(+19.66)} & 82.51{\tiny\color{blue}(+16.88)} & 94.40{\tiny\color{blue}(+16.98)} \\
        CRAFT$_{\text{v3}}$ & 51.86{\tiny\color{blue}(+12.75)} & 62.37{\tiny\color{blue}(+12.87)} & 84.60{\tiny\color{blue}(+16.13)} & 63.81{\tiny\color{blue}(+5.43)} & 78.55{\tiny\color{blue}(+6.64)} & 96.23{\tiny\color{blue}(+5.70)} & 76.62{\tiny\color{blue}(+15.55)} & 83.23{\tiny\color{blue}(+13.38)} & 95.97{\tiny\color{blue}(+11.45)} \\
        CRAFT$_{\text{v4}}$ & 48.02{\tiny\color{blue}(+9.61)} & 59.94{\tiny\color{blue}(+10.80)} & 83.17{\tiny\color{blue}(+6.64)} & 62.64{\tiny\color{blue}(+2.52)} & 77.84{\tiny\color{blue}(+3.44)} & 94.96{\tiny\color{blue}(+1.24)} & 75.60{\tiny\color{blue}(+14.80)} & 82.04{\tiny\color{blue}(+11.67)} & 92.90{\tiny\color{blue}(+6.25)} \\
        CRAFT$_{\text{v5}}$ & 49.95{\tiny\color{blue}(+18.52)} & 60.51{\tiny\color{blue}(+18.60)} & - & 59.80{\tiny\color{blue}(+6.42)} & 74.79{\tiny\color{blue}(+6.82)} & - & 60.34{\tiny\color{blue}(+9.33)} & 67.72{\tiny\color{blue}(+8.51)} & - \\
        \bottomrule
    \end{tabular*}
    \caption{7B comparison across trace variants (\textsc{CRAFT}$_{\text{v1}}$--\textsc{CRAFT}$_{\text{v5}}$) for Base, SFT, w/o judge reward, and full CRAFT. Faithfulness is the binary overall-consistency pass rate. Colored deltas are relative to Base. Because \textsc{CRAFT}$_{\text{v5}}$ is answer-only, Faithfulness is not applicable, and its w/o-judge and full results are identical.}
    \label{tab:comparison_7b_v1_v5}
\end{table*}

\noindent\textbf{Why no direct supervision for Plan/Reason?}
Unlike $\mathcal{E}$ (discrete indices) and $a$ (short spans), $\pi$ and $\rho$ are open-ended generations with high semantic variance, making direct supervision $\min_\theta \mathcal{L}(\pi_\theta, \pi^*)$ via teacher traces brittle and prone to format collapse. Instead, we supervise \textbf{functional correctness}: $\pi$ is valid iff $\rho$ follows it ($r_{\pi\rightarrow\rho}=1$); $\rho$ is valid iff it is grounded ($r_{\mathrm{ground}}=1$). This lets the model discover optimal reasoning paths within the auditable structure without imitating a fixed trace.

\noindent\textbf{Method Scope.}
CRAFT assumes that $D$ has already been retrieved and does not update the retriever, search policy, or evidence ordering. It optimizes only structured post-retrieval generation. Faithfulness here is an operational property of the emitted trace---internal consistency and support from $D$---rather than a claim that the trace recovers the model's latent causal computation. CRAFT therefore complements retrieval-trajectory optimization and post-hoc faithfulness evaluation.

\section{Experiments}
\label{sec:experiments}

\subsection{Experimental Setup}

\noindent\textbf{Datasets.}
We evaluate CRAFT on three multi-hop QA benchmarks: HotpotQA~\cite{yang2018hotpotqa}, 2WikiMultiHopQA~\cite{ho2020constructing}, and MuSiQue~\cite{trivedi2022musique}. These datasets represent diverse multi-hop reasoning types including bridge reasoning, comparison, and compositional questions. For training, we construct a mixed dataset of 20,000 entries sampled from all three benchmarks. For each dataset, we randomly select 1,000 test examples and evaluate them ten times.

\noindent\textbf{Baselines.}
We compare against: (1) \textbf{Open-source models}: Qwen2.5 series (0.5B, 1.5B, 3B, 7B)~\cite{qwen2.5} and Qwen3 series (4B, 30B-A3B)~\cite{qwen3} via in-context prompting; (2) \textbf{Closed-source API models}: GPT-5-mini~\cite{openai2025gpt5}, DeepSeek-V3.2~\cite{liu2025deepseek}, and Gemini-2.5-Flash~\cite{google2025gemini25flash}. We train CRAFT at four scales (0.5B, 1.5B, 3B, 7B) using both SFT and GRPO.

\noindent\textbf{Metrics.}
We report \textbf{Exact Match (EM)}, token-level \textbf{F1}, and \textbf{Faithfulness}. During training, $R_{\mathrm{faith}}$ averages the applicable binary component checks; reported Faithfulness instead uses the pass rate of a separate binary overall-consistency verdict (Algorithm~\ref{alg:judge_v1}).

\noindent\textbf{Implementation Details.}
Model, hardware, judge, reward-weight, and aggregation details are provided in Appendix~\ref{sec:impl_details}.

\subsection{Main Results}

\noindent\textbf{Gains start at 1.5B; 0.5B is trace-sensitive.}
For 7B, \textsc{CRAFT} improves EM over base Qwen2.5-7B by 17.30/7.59/20.36 points on MuSiQue/HotpotQA/2WikiMHQA, with relative gains of 48.3\%/13.4\%/35.2\%. The 1.5B and 3B models also improve on all datasets under \textsc{CRAFT}$_{\text{v1}}$. The 0.5B model stays near the accuracy floor under this complex trace (0.00\%/0.44\%/1.35\% EM), indicating the benefit requires sufficient capacity or a simpler template.

\noindent\textbf{Faithfulness improves under full CRAFT.}
Full CRAFT-7B reaches 83.87\%/96.31\%/96.79\% Faithfulness on MuSiQue/HotpotQA/2WikiMHQA, improving over Base by 25.40/13.01/19.87 points and over SFT by 32.65/12.94/17.79 points. Under \textsc{CRAFT}$_{\text{v1}}$, the 1.5B, 3B, and 7B trained models exceed Base on all three datasets. The 0.5B model remains near the faithfulness floor, consistent with its difficulty following the full trace format.

\noindent\textbf{\textsc{CRAFT}$_{\text{7B}}$ reaches API-level answer performance.}
\textsc{CRAFT}$_{\text{7B}}$ achieves 53.13\% EM on MuSiQue, compared with the best API EM of 53.70\%. On HotpotQA it reaches 64.06\%/78.86\% EM/F1, compared with the best API EM/F1 scores of 63.60\% and 80.22\%, respectively; on 2WikiMHQA it reaches 78.23\%/84.18\%, compared with 70.47\%/81.19\%. \textsc{CRAFT}$_{\text{7B}}$ matches or exceeds the best API EM on two of three benchmarks with strong Faithfulness.

\begin{figure*}[t]
    \centering
    \includegraphics[width=\textwidth]{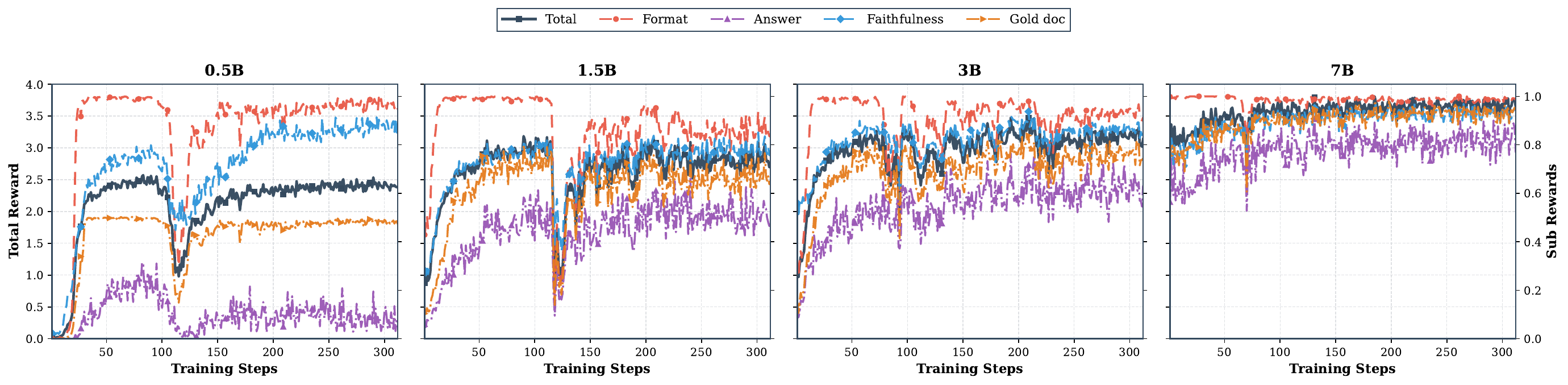}
    \caption{Training dynamics of GRPO across scales (0.5B, 1.5B, 3B, 7B) using \textsc{CRAFT}$_{\text{v1}}$. Left y-axis: total reward; right y-axis: component rewards ($R_{\mathrm{fmt}}$, $R_{\mathrm{gold}}$, $R_{\mathrm{faith}}$, $R_{\mathrm{ans}}$).}
    \label{fig:training_dynamics}
\end{figure*}

\subsection{Ablation Study}
\label{sec:ablation}

We conduct two analyses: (1) a trace-template ablation with aligned deterministic rewards, and (2) a judge-reward ablation. Starting from the full template \textsc{CRAFT}$_{\text{v1}}$ (\texttt{<plan>}+\texttt{<gold\_docs>}+\texttt{<reason>}+\texttt{<answer>}), we progressively remove components:
\textsc{CRAFT}$_{\text{v2}}$ removes \texttt{<plan>},
\textsc{CRAFT}$_{\text{v3}}$ removes \texttt{<gold\_docs>},
\textsc{CRAFT}$_{\text{v4}}$ retains \texttt{<reason>}+\texttt{<answer>},
and \textsc{CRAFT}$_{\text{v5}}$ retains only \texttt{<answer>}.
Table~\ref{tab:comparison_7b_v1_v5} reports Qwen2.5-7B results across Base, SFT, w/o judge reward, and full CRAFT. For \textsc{CRAFT}$_{\text{v1}}$--\textsc{CRAFT}$_{\text{v4}}$, the w/o-judge condition removes $R_{\mathrm{faith}}$ while retaining the other applicable rewards and training settings. The two \textsc{CRAFT}$_{\text{v5}}$ GRPO rows are identical because this answer-only variant has no judge-reward term.

\noindent\textbf{Trace structure has dataset-dependent effects.}
Answer-only \textsc{CRAFT}$_{\text{v5}}$ achieves the lowest base EM (31.43\%/53.38\%/51.01\%), underperforming \textsc{CRAFT}$_{\text{v4}}$ by 6.98/6.74/9.79 points on MuSiQue/HotpotQA/2WikiMHQA. Under GRPO, the strongest structured variants reach 53.13\%/65.28\%/78.23\% EM, exceeding v5 by 3.18/5.48/17.89 points on the three datasets. Thus, structured traces remain beneficial, especially on 2WikiMHQA.

\noindent\textbf{Simpler traces achieve higher base faithfulness but lower auditability.}
For base models, simpler \textsc{CRAFT}$_{\text{v4}}$ yields higher Faithfulness (76.53\%/93.72\%/86.65\%) than full \textsc{CRAFT}$_{\text{v1}}$ (58.47\%/83.30\%/76.92\%), because the judge assesses fewer cross-field relations. After GRPO, template effects remain dataset-dependent, while \textsc{CRAFT}$_{\text{v1}}$/\textsc{CRAFT}$_{\text{v2}}$ retain greater auditability through explicit planning and citation.

\noindent\textbf{SFT gives inconsistent faithfulness.}
Across \textsc{CRAFT}$_{\text{v1}}$--\textsc{CRAFT}$_{\text{v4}}$ and all three datasets, full CRAFT improves over Base by 2.52--20.36 EM points and 1.24--25.40 Faithfulness points. SFT is less consistent: it reduces EM in 5 of 12 cells and Faithfulness in 8 of 12 cells, with maximum drops of 8.30 and 10.73 points, respectively.

\noindent\textbf{Judge reward acts as a semantic regularizer.}
At 7B, full CRAFT outperforms the w/o-judge condition in all 12 applicable cells. The Full-minus-w/o gains span 1.57--2.02 points for EM, 1.50--2.41 for F1, and 1.49--2.31 for Faithfulness, with respective averages of 1.75, 1.89, and 1.89 points. The gains appear in Faithfulness, EM, and F1, suggesting that judge feedback does more than optimize trace style: it discourages unsupported reasoning paths and helps preserve answer quality while making the derivation more auditable.

\subsection{Training Dynamics}
\label{sec:training_dynamics}

\noindent\textbf{Training reward scales with capacity.}
After 312 GRPO steps, total rewards for \textsc{CRAFT}$_{\text{0.5B}}$, \textsc{CRAFT}$_{\text{1.5B}}$, \textsc{CRAFT}$_{\text{3B}}$, and \textsc{CRAFT}$_{\text{7B}}$ are 2.40/2.84/3.18/3.67, answer rewards are 0.07/0.51/0.62/0.82, and citation rewards are 0.48/0.67/0.76/0.93. Their monotonic increases show that larger models better convert structured RL feedback into accurate, auditable traces.

\noindent\textbf{Online rewards support evaluation Faithfulness.}
After 312 steps, format rewards remain high (0.97/0.86/0.93/0.99), and mean per-check judge rewards reach 0.88/0.78/0.86/0.93. These online signals show that trained policies learn both schema compliance and judge-supported reasoning, complementing Table~\ref{tab:main_results_v1}'s offline binary Faithfulness on evaluation traces from frozen checkpoints.

\noindent\textbf{Figure~\ref{fig:training_dynamics} reveals a capacity-dependent transition.}
All scales acquire structural rewards and recover from transient drops. Answer reward remains low for \textsc{CRAFT}$_{\text{0.5B}}$, rises for \textsc{CRAFT}$_{\text{1.5B}}$ and \textsc{CRAFT}$_{\text{3B}}$, and peaks for \textsc{CRAFT}$_{\text{7B}}$; its balanced profile indicates joint optimization beyond schema compliance. This pattern separates schema learning from semantic reasoning: small models can satisfy the output form before they can reliably use it for answer derivation, whereas larger models turn the same structured reward into coordinated gains in citation, answer, and faithfulness.
\begin{table}[t]
    \centering
    \scriptsize
    \setlength{\tabcolsep}{2pt}
    \begin{tabular*}{\columnwidth}{@{\extracolsep{\fill}}lccccccc@{}}
        \toprule
        \textbf{Aspect} & \textbf{Both} & \textbf{Both} & \textbf{LLM\textcolor{green!50!black}{$\checkmark$}} & \textbf{LLM\textcolor{red}{$\times$}} & \textbf{Agr.} & \textbf{$\kappa$} \\
         & \textbf{Pass} & \textbf{Fail} & \textbf{Human\textcolor{red}{$\times$}} & \textbf{Human\textcolor{green!50!black}{$\checkmark$}} & & \\
        \midrule
        Plan-Reason & 393 & 78 & 15 & 14 & 94.2\% & 0.81 \\
        Evidence Gr. & 375 & 83 & 25 & 17 & 91.6\% & 0.75 \\
        Answer Deriv. & 410 & 71 & 9 & 10 & 96.2\% & 0.86 \\
        Gold Doc Cit. & 370 & 80 & 33 & 17 & 90.0\% & 0.70 \\
        \midrule
        \textbf{Macro Avg.} & \textbf{387.0} & \textbf{78.0} & \textbf{20.5} & \textbf{14.5} & \textbf{93.0\%} & \textbf{0.78} \\
        \bottomrule
    \end{tabular*}
    \caption{Human validation on 500 MuSiQue samples: LLM-human confusion counts, agreement, and Cohen's $\kappa$. Macro Avg. is the unweighted average across audit dimensions.}
    \label{tab:human_validation}
\end{table}

\subsection{Human Validation of Judge Reliability}
\label{sec:human_validation}

Two NLP annotators independently assessed 500 randomly sampled MuSiQue reasoning traces from \textsc{CRAFT}$_{\text{7B}}$ using the same four-aspect binary criteria as the LLM judge (Algorithm~\ref{alg:judge_v1}). Across the four audit dimensions, the judge reaches \textbf{93.0\%} macro-averaged observed agreement with human labels and a mean Cohen's $\kappa$ of $0.78$~\cite{artstein2017inter}, indicating substantial agreement (Table~\ref{tab:human_validation}). Agreement is strongest for Answer Derivation ($\kappa = 0.86$) and lowest for Gold Document Citation ($\kappa = 0.70$), the most ambiguity-prone dimension. These results show that the judge aligns well with human audit decisions, reducing but not eliminating the circularity concern from using Qwen3-30B-A3B for both training and primary evaluation~\cite{chen2024humans,gu2024survey}.

\section{Conclusion}
\label{sec:conclusion}
We presented \textsc{CRAFT}, a GRPO framework for structured, machine-auditable post-retrieval reasoning. The central finding is that \textsc{CRAFT} unlocks task-specific reasoning capacity without trading accuracy for faithfulness: gains emerge from 1.5B upward, whereas \textsc{CRAFT}$_{\text{0.5B}}$ benefits from simpler traces. At 7B, \textsc{CRAFT} reaches API-level answer performance, and the judge-reward ablation shows consistent gains in both answer quality and Faithfulness. Template ablations further reveal a capacity-dependent trade-off between auditability and learnability, while human validation supports the judge's reliability. More broadly, structured RL unlocks accurate, auditable task-specific reasoning.

\clearpage
\section*{Acknowledgments}
This research is supported by the National Key R\&D Program of China (No.\ 2023YFC3303800). This paper is supported by the National Natural Science Foundation of China (No. 62406161)

\section*{Limitations}

Our human validation of judge reliability is conducted on 500 samples from MuSiQue only. While we select MuSiQue as the most challenging benchmark, extending human annotation to all three datasets would further strengthen the reliability analysis; we leave this to future work due to the cost of expert annotation on multi-hop reasoning traces. Second, due to the substantial computational cost of GRPO training with online judge inference, our experiments are limited to models up to 7B parameters trained on 20K samples from three multi-hop QA benchmarks (HotpotQA, 2WikiMHQA, MuSiQue). The observed relationship between Faithfulness and answer accuracy is template- and dataset-dependent. Scaling to larger models remains important future work.

\bibliography{ref}

\clearpage
\appendix

\section{Implementation Details}
\label{sec:impl_details}
\begin{table}[h]
    \centering
    \scriptsize
    \setlength{\tabcolsep}{2pt}
    \begin{tabular*}{\columnwidth}{@{\extracolsep{\fill}}lcc@{}}
        \toprule
        \textbf{Hyperparameter} & \textbf{0.5B--3B} & \textbf{7B} \\
        \midrule
        Per-device batch & 8 & 8 \\
        Training GPUs & 4 & 4 \\
        Generations & 8 & 8 \\
        Gradient accumulation & 16 & 16 \\
        Effective completions/update & 512 & 512 \\
        Prompts/update & 64 & 64 \\
        Training steps & 312 & 312 \\
        Max completion length & 256 & 512 \\
        Model/vLLM max length & 8,192/8,192 & 8,192/8,192 \\
        Learning rate & 5e-6 & 3e-6 \\
        LR scheduler & Cosine & Cosine \\
        Warmup ratio & 0.1 & 0.1 \\
        GRPO clip $\varepsilon$ & 0.2 & 0.2 \\
        KL coefficient $\beta$ & 0.04 & 0.04 \\
        Reward weights & 1.0 each & 1.0 each \\
        Precision & bfloat16 & bfloat16 \\
        Attention & Flash Attention 2 & Flash Attention 2 \\
        \bottomrule
    \end{tabular*}
    \caption{Key GRPO training hyperparameters.}
    \label{tab:training_config}
\end{table}
This section expands on the experimental setup described in \S\ref{sec:experiments}.

\noindent\textbf{Training Framework.}
We implement CRAFT using \textbf{MS-Swift}~\cite{zhao2024swiftascalablelightweightinfrastructure}, an efficient fine-tuning framework for large language models. MS-Swift provides native support for GRPO training with vLLM acceleration, enabling efficient on-policy sampling during reinforcement learning. All CRAFT GRPO runs use four NVIDIA H100 96GB GPUs for policy training; the remaining four GPUs on the 8-GPU node host the local judge. Table~\ref{tab:training_config} summarizes the key hyperparameters.

\noindent\textbf{Model Configuration.}
All CRAFT policy models are initialized from the Qwen2.5-Instruct series~\cite{qwen2.5}; we train four scales: 0.5B, 1.5B, 3B, and 7B parameters. All models use bfloat16 precision with Flash Attention 2 for memory-efficient training. For SFT baselines, we fine-tune each model on the same training data using standard cross-entropy loss with a learning rate of 2e-5 and 3 epochs.

\noindent\textbf{Judge Model.}
All primary judge calls in the reported experiments use \textbf{Qwen3-30B-A3B}~\cite{qwen3}: this includes online calls that produce $R_{\mathrm{faith}}$ during GRPO and offline calls that compute the main-table Faithfulness metric. Training and evaluation nevertheless use separate data flows. During training, the judge scores newly sampled policy completions and the mean of the applicable binary checks enters the GRPO reward, providing graded feedback. During evaluation, a frozen checkpoint first generates traces for benchmark evaluation samples; Qwen3-30B-A3B then re-judges those saved traces at temperature 0.0, without reusing training rewards or training traces. Every evaluation trace receives one binary \texttt{overall\_consistency} verdict, and dataset-level Faithfulness is the pass rate over all evaluation traces.

\noindent\textbf{Evaluation Aggregation.}
For each benchmark, we report the mean over ten 1,000-example evaluation runs. The same method--variant--dataset aggregate is reused wherever it appears in multiple tables.

\noindent\textbf{Reward Function Configuration.}
The active reward components vary across trace templates based on the available structural fields:
\textbf{CRAFT$_{\text{v1}}$/CRAFT$_{\text{v2}}$}: $R_{\mathrm{fmt}}$, $R_{\mathrm{ans}}$, $R_{\mathrm{gold}}$, $R_{\mathrm{faith}}$ (full supervision);
\textbf{CRAFT$_{\text{v3}}$/CRAFT$_{\text{v4}}$}: $R_{\mathrm{fmt}}$, $R_{\mathrm{ans}}$, $R_{\mathrm{faith}}$ (no citation reward as \texttt{<gold\_docs>} is absent);
\textbf{CRAFT$_{\text{v5}}$}: $R_{\mathrm{fmt}}$, $R_{\mathrm{ans}}$ only (no reasoning fields for faithfulness audit).
All reward weights are set to 1.0. Each component lies in $[0,1]$: $R_{\mathrm{fmt}}$ and $R_{\mathrm{ans}}$ are binary, $R_{\mathrm{gold}}$ is a three-level overlap score, and $R_{\mathrm{faith}}$ is an average of binary checks. Equal weighting therefore keeps the nominal reward ranges comparable, but does not imply equal variance or equal influence on optimization. $R_{\mathrm{fmt}}$ rises rapidly in most runs but remains active whenever sampled outputs violate the schema (Figure~\ref{fig:training_dynamics}). In the w/o-judge condition in Table~\ref{tab:comparison_7b_v1_v5}, $R_{\mathrm{faith}}$ is removed while all other applicable reward components are retained.

\begin{figure}[t]
    \centering
    \includegraphics[width=0.9\linewidth]{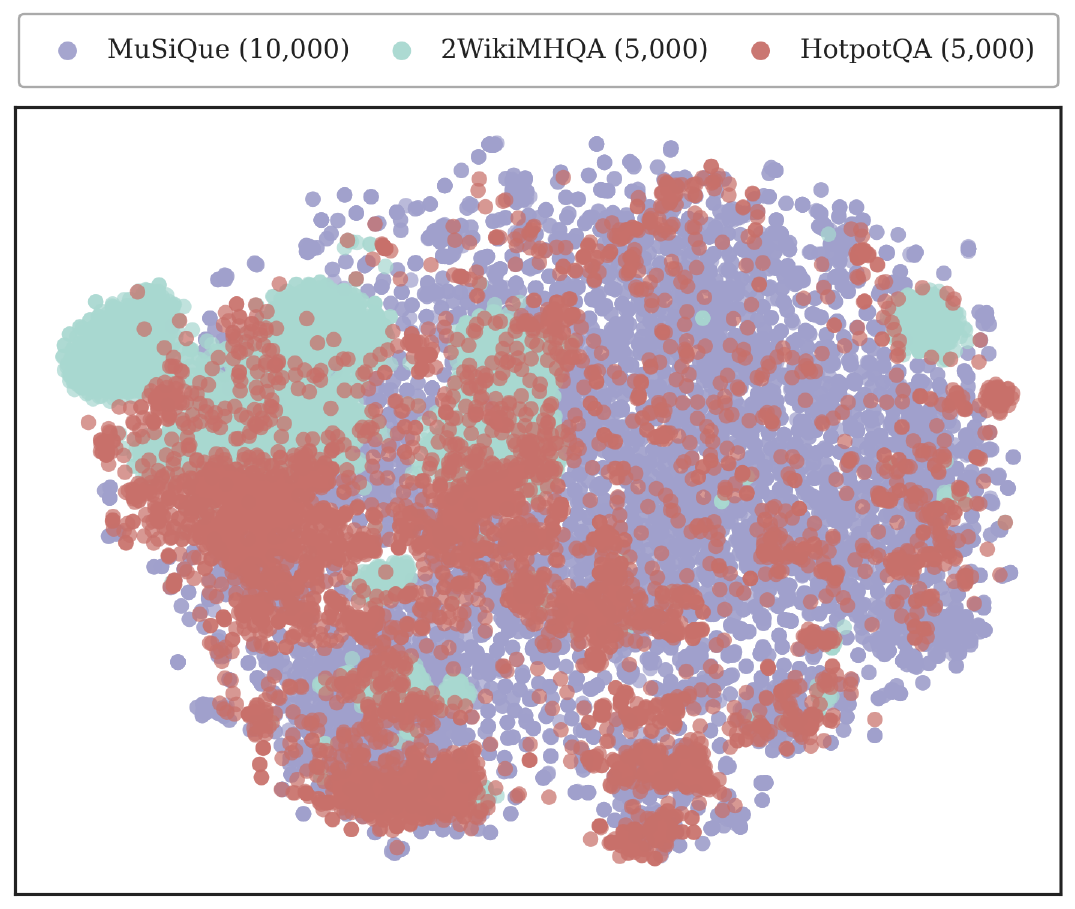}
    \caption{Qualitative t-SNE projection of sentence-encoder query embeddings for the 20,000 training samples: 10,000 MuSiQue (lavender blue), 5,000 HotpotQA (coral red), and 5,000 2WikiMHQA (mint green).} 
    \label{fig:data_distribution}
\end{figure}

\noindent\textbf{Training Data.}
Each prompt template uses a dedicated training dataset of 20,000 samples: 10,000 MuSiQue, 5,000 HotpotQA, and 5,000 2WikiMHQA. Figure~\ref{fig:data_distribution} provides a qualitative t-SNE projection of sentence-encoder query embeddings; it is included as a distribution diagnostic rather than a quantitative measure of dataset separation.

\section{Additional Results}
\label{sec:additional_results}

This section provides supplementary results that expand on the main experiments in \S\ref{sec:experiments}, including out-of-distribution evaluation, efficiency and cost analysis, training dynamics, full variant results, format analysis, and capacity-dependent template preference.

\subsection{Out-of-Distribution Evaluation}
\label{sec:app_ood}

We evaluate \textsc{CRAFT} on MultiHopRAG~\cite{tang2024multihoprag}, a held-out benchmark unseen during training and drawn from a distinct news corpus with different domains and question types. Because the official release does not provide a gold-plus-distractor setting, we construct each input from up to five gold documents and corpus-sampled distractors to total 10 documents (seed 42), using the same \textsc{CRAFT}$_{\text{v1}}$ prompt format.

\begin{table}[t]
    \centering
    \scriptsize
    \setlength{\tabcolsep}{3pt}
    \begin{tabular*}{\columnwidth}{@{\extracolsep{\fill}}llrrr@{}}
        \toprule
        \textbf{Model} & \textbf{Dataset} & \textbf{EM} & \textbf{F1} & \textbf{Faith.} \\
        \midrule
        Qwen2.5-7B$_{\text{Base}}$ & MultiHopRAG & 56.59 & 57.64 & 19.02 \\
        \textsc{CRAFT}$_{\text{7B}}$ & MultiHopRAG & \textbf{63.45} & \textbf{64.82} & \textbf{48.73} \\
        \midrule
        $\Delta$ & -- & +6.86 & +7.18 & +29.71 \\
        \bottomrule
    \end{tabular*}
    \caption{Out-of-distribution evaluation (\%) on MultiHopRAG. Both models use the same \textsc{CRAFT}$_{\text{v1}}$ prompt and 10-document input construction.}
    \label{tab:ood_multihoprag}
\end{table}

\textsc{CRAFT}$_{\text{7B}}$ improves over Base by 6.86 EM, 7.18 F1, and 29.71 Faithfulness points. These gains show that structured traces and faithfulness supervision transfer beyond the training distribution.

\subsection{Efficiency and Cost Analysis}
\label{sec:app_efficiency}

\subsubsection{Judge-Inference Overhead}

We conduct a controlled 7B comparison that holds the configuration, data, hardware, and training length fixed, with both runs completing 312 GRPO steps and differing only in whether $R_{\mathrm{faith}}$ is active.

\begin{table}[t]
    \centering
    \small
    \begin{tabular*}{\columnwidth}{@{\extracolsep{\fill}}lrr@{}}
        \toprule
        \textbf{Configuration} & \textbf{Time/step (s)} & \textbf{Total (h)} \\
        \midrule
        \textsc{CRAFT}$_{\text{7B}}$ w/o judge & 110.5 & 9.58 \\
        \textsc{CRAFT}$_{\text{7B}}$ w/ judge & 135.5 & 11.74 \\
        \midrule
        $\Delta$ overhead & +25.0 & +2.16 \\
        \bottomrule
    \end{tabular*}
    \caption{Controlled 7B judge-overhead comparison over 312 GRPO steps.}
    \label{tab:judge_overhead}
\end{table}

Judge inference adds 22.6\% training overhead, equivalent to 2\,h\,10\,min over 312 steps. Per-call judge latency is 5--7\,s, but much of this cost is absorbed by co-located scheduling because policy computation dominates at 7B. For context, offline rejection-sampling CRAFT (RS-CRAFT) takes approximately 8\,h end-to-end, including 5\,h\,50\,min of judge scoring; online \textsc{CRAFT}$_{\text{7B}}$ requires approximately 3\,h\,44\,min more wall-clock time.

\subsubsection{Trace Length and Generation Efficiency}

Trace length varies by design across \textsc{CRAFT}$_{\text{v1}}$--\textsc{CRAFT}$_{\text{v5}}$, exposing an explicit auditability--efficiency trade-off. Table~\ref{tab:trace_efficiency} averages output-only token counts across the three datasets for the 7B variants.

\begin{table}[t]
    \centering
    \scriptsize
    \setlength{\tabcolsep}{4pt}
    \begin{tabular*}{\columnwidth}{@{\extracolsep{\fill}}lr@{}}
        \toprule
        \textbf{Variant} & \textbf{Avg. output tokens} \\
        \midrule
        \textsc{CRAFT}$_{\text{v1}}$ & 144.6 \\
        \textsc{CRAFT}$_{\text{v2}}$ & 88.5 \\
        \textsc{CRAFT}$_{\text{v3}}$ & 152.5 \\
        \textsc{CRAFT}$_{\text{v4}}$ & 97.5 \\
        \textsc{CRAFT}$_{\text{v5}}$ & \textbf{11.3} \\
        \bottomrule
    \end{tabular*}
    \caption{Average 7B output length across MuSiQue, HotpotQA, and 2WikiMHQA. Token counts exclude input tokens.}
    \label{tab:trace_efficiency}
\end{table}

\textsc{CRAFT}$_{\text{v2}}$ and \textsc{CRAFT}$_{\text{v4}}$ provide compact structured operating points at 88.5 and 97.5 tokens, respectively. The full v1 trace provides the richest audit structure at a higher output cost, while answer-only v5 is shortest but removes all intermediate structure. This spectrum lets practitioners select a variant according to generation budget and required trace structure.

\begin{figure*}[t]
    \centering
    \includegraphics[width=\textwidth]{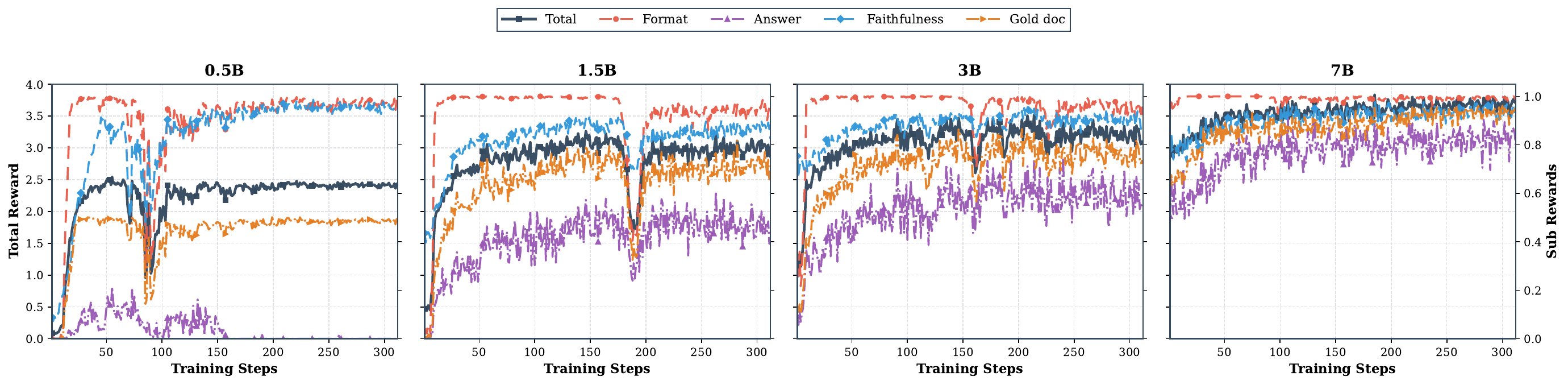}
    \caption{Training dynamics using \textsc{CRAFT}$_{\text{v2}}$ template.}
    \label{fig:training_v2}
\end{figure*}

\begin{figure*}[ht!]
    \centering
    \includegraphics[width=\textwidth]{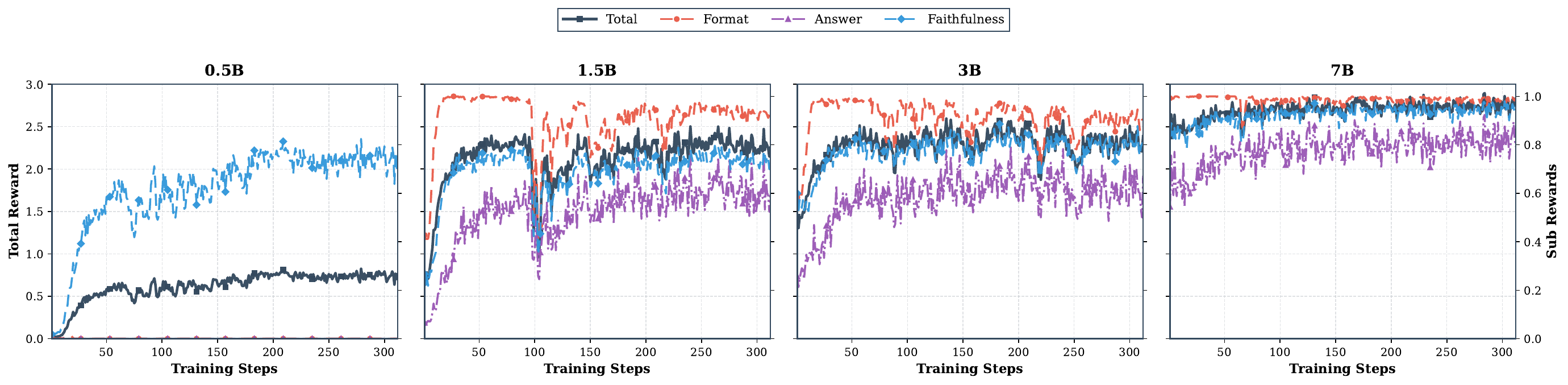}
    \caption{Training dynamics using \textsc{CRAFT}$_{\text{v3}}$ template. Note: $R_{\mathrm{gold}}$ not applicable.}
    \label{fig:training_v3}
\end{figure*}

\begin{figure*}[ht!]
    \centering
    \includegraphics[width=\textwidth]{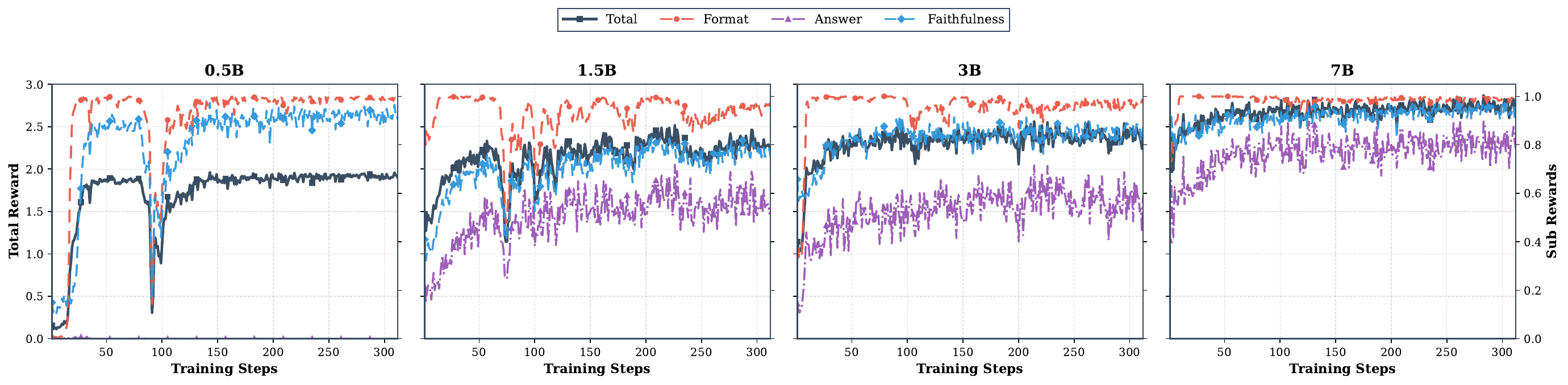}
    \caption{Training dynamics using \textsc{CRAFT}$_{\text{v4}}$ template. Note: $R_{\mathrm{gold}}$ not applicable.}
    \label{fig:training_v4}
\end{figure*}

\begin{figure*}[ht!]
    \centering
    \includegraphics[width=\textwidth]{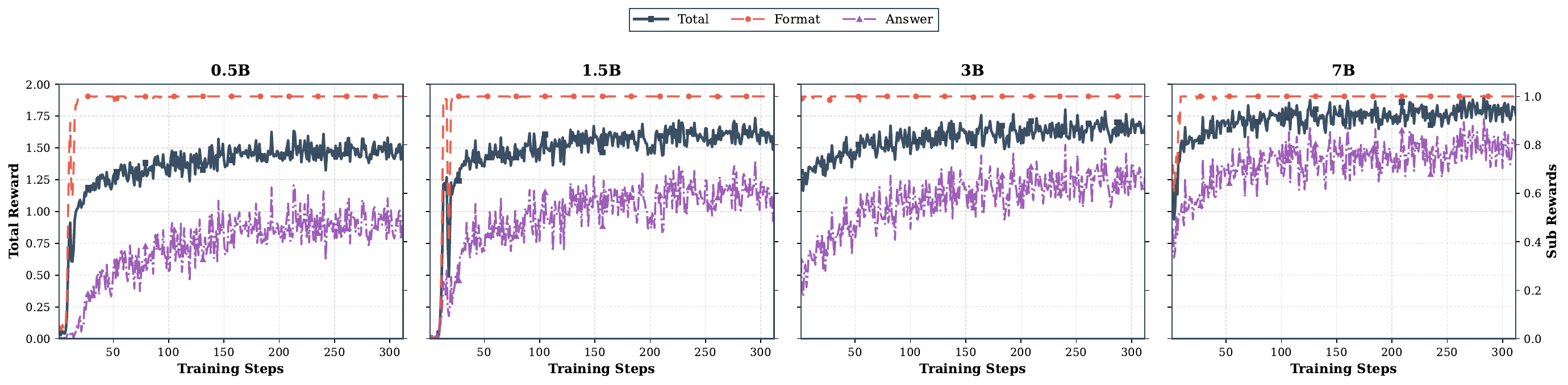}
    \caption{Training dynamics using \textsc{CRAFT}$_{\text{v5}}$ template. Note: $R_{\mathrm{faith}}$ and $R_{\mathrm{gold}}$ not applicable.}
    \label{fig:training_v5}
\end{figure*}

\subsection{Training Dynamics for Additional Variants}
\label{sec:app_training}

Figures~\ref{fig:training_v2}--\ref{fig:training_v5} show the reward trajectories of \textsc{CRAFT}$_{\text{v2}}$--\textsc{CRAFT}$_{\text{v5}}$ throughout their complete 312-step training runs. The runs recover from transient dips and settle into stable regimes. For the structured variants, total reward generally increases with model scale; under the answer-only \textsc{CRAFT}$_{\text{v5}}$, the 3B and 7B runs reach similar end-of-training levels.

\textbf{Template-specific patterns}: (1) \textsc{CRAFT}$_{\text{v2}}$ shows dynamics similar to \textsc{CRAFT}$_{\text{v1}}$, with all four reward components contributing. (2) \textsc{CRAFT}$_{\text{v3}}$ and \textsc{CRAFT}$_{\text{v4}}$ omit $R_{\mathrm{gold}}$, while their larger-scale runs retain stable reward growth. (3) \textsc{CRAFT}$_{\text{v5}}$ uses only $R_{\mathrm{fmt}}$ and $R_{\mathrm{ans}}$ and is particularly effective at 0.5B, consistent with the capacity-dependent template preference in Table~\ref{tab:format_accuracy}.

\begin{table*}[t]
    \centering
    \scriptsize
    \setlength{\tabcolsep}{3pt}
    \begin{tabular*}{\textwidth}{@{\extracolsep{\fill}}lcccccccccc@{}}
        \toprule
        \textbf{Method} & \textbf{Type} & \multicolumn{3}{c}{\textbf{MuSiQue}} & \multicolumn{3}{c}{\textbf{HotpotQA}} & \multicolumn{3}{c}{\textbf{2WikiMHQA}} \\
         &  & EM & F1 & Faith. & EM & F1 & Faith. & EM & F1 & Faith. \\
        \midrule
        \rowcolor{gray!20}\multicolumn{11}{l}{\textit{Open-source}} \\
        \quad Qwen2.5-0.5B & Pre-Trained & 0.00 & 1.03 & 1.51 & 1.30 & 3.99 & 4.28 & 1.76 & 5.02 & 1.98 \\
        \quad Qwen2.5-1.5B & Pre-Trained & 4.01 & 7.52 & 6.83 & 17.18 & 25.28 & 21.20 & 15.49 & 21.57 & 18.29 \\
        \quad Qwen2.5-3B & Pre-Trained & 22.61 & 31.63 & 42.11 & 49.01 & 63.94 & 74.22 & 44.24 & 52.68 & 58.64 \\
        \quad Qwen2.5-7B & Pre-Trained & 34.24 & 44.77 & 66.15 & 54.56 & 69.62 & 87.80 & 55.60 & 65.63 & 77.42 \\
        \quad Qwen3-4B & Pre-Trained & 34.42 & 43.48 & 57.03 & 60.24 & 75.10 & 87.90 & 63.78 & 72.82 & 82.02 \\
        \quad Qwen3-30B-A3B & Pre-Trained & 48.30 & 57.65 & 71.53 & 63.20 & 78.23 & 89.78 & 69.59 & 78.25 & 88.20 \\
        \midrule
        \rowcolor{gray!20}\multicolumn{11}{l}{\textit{API}} \\
        \quad GPT-5-mini & API & \uwave{52.34} & 66.91 & \uwave{87.96} & 62.01 & \uwave{80.14} & \uwave{96.75} & 69.47 & \uwave{80.47} & 94.41 \\
        \quad DeepSeek-V3.2 & API & 51.63 & \uwave{67.70} & 85.90 & 61.57 & 79.38 & 96.47 & \uwave{70.13} & 78.94 & \uwave{97.10} \\
        \quad Gemini-2.5-Flash & API & 50.92 & 61.88 & 82.07 & \uwave{63.26} & 79.00 & 94.54 & 69.66 & 79.40 & 94.16 \\
        \midrule
        \rowcolor{gray!20}\multicolumn{11}{l}{\textit{SFT}} \\
        \quad Qwen2.5-0.5B & Trained & 0.00 & 0.61 & 0.80 & 0.42 & 2.16 & 2.48 & 1.09 & 2.83 & 1.08 \\
        \quad Qwen2.5-1.5B & Trained & 3.39 & 7.03 & 9.08 & 18.17 & 26.97 & 37.48 & 28.80 & 34.46 & 37.74 \\
        \quad Qwen2.5-3B & Trained & 20.85 & 31.22 & 37.55 & 50.09 & 63.94 & 79.68 & 53.59 & 62.89 & 73.32 \\
        \quad Qwen2.5-7B & Trained & 28.38 & 40.32 & 56.00 & 55.51 & 70.42 & 88.14 & 59.67 & 67.86 & 76.31 \\
        \midrule
        \rowcolor{gray!20}\multicolumn{11}{l}{\textit{Ours (RL)}} \\
        \quad CRAFT$_{\text{0.5B}}$ & Trained & 0.00 & 2.94 & \underline{90.76} & 0.00 & 6.18 & 83.78 & 0.00 & 7.28 & 89.38 \\
        \quad CRAFT$_{\text{1.5B}}$ & Trained & 19.43 & 24.90 & 56.44 & 48.65 & 61.90 & 72.14 & 43.37 & 49.22 & 50.49 \\
        \quad CRAFT$_{\text{3B}}$ & Trained & 37.38 & 45.85 & 72.16 & 57.86 & 73.41 & 87.14 & 55.89 & 62.18 & 65.66 \\
        \quad CRAFT$_{\text{7B}}$ & Trained & \underline{51.31} & \underline{63.06} & 82.49 & \underline{65.28} & \underline{80.10} & \underline{96.38} & \underline{75.26} & \underline{82.51} & \underline{94.40} \\
        \bottomrule
    \end{tabular*}
    \caption{Main results for \textsc{CRAFT}$_{\text{v2}}$ across three multi-hop QA benchmarks. This variant removes the planning field while retaining explicit document citation, reasoning, and answer extraction. EM/F1 and binary overall-consistency Faithfulness are reported in percentage. \uwave{Wavy underline}: best among API models; \underline{straight underline}: best among all other models.}
    \label{tab:main_results_v2}
\end{table*}

\begin{table*}[ht!]
    \centering
    \scriptsize
    \setlength{\tabcolsep}{3pt}
    \begin{tabular*}{\textwidth}{@{\extracolsep{\fill}}lcccccccccc@{}}
        \toprule
        \textbf{Method} & \textbf{Type} & \multicolumn{3}{c}{\textbf{MuSiQue}} & \multicolumn{3}{c}{\textbf{HotpotQA}} & \multicolumn{3}{c}{\textbf{2WikiMHQA}} \\
         &  & EM & F1 & Faith. & EM & F1 & Faith. & EM & F1 & Faith. \\
        \midrule
        \rowcolor{gray!20}\multicolumn{11}{l}{\textit{Open-source}} \\
        \quad Qwen2.5-0.5B & Pre-Trained & 0.21 & 1.56 & 0.88 & 2.05 & 7.36 & 6.29 & 1.74 & 8.74 & 4.42 \\
        \quad Qwen2.5-1.5B & Pre-Trained & 2.35 & 6.83 & 15.25 & 7.27 & 16.56 & 34.59 & 9.60 & 16.95 & 21.61 \\
        \quad Qwen2.5-3B & Pre-Trained & 17.43 & 26.29 & 38.53 & 38.49 & 51.59 & 70.02 & 34.07 & 43.98 & 61.71 \\
        \quad Qwen2.5-7B & Pre-Trained & 39.11 & 49.50 & 68.47 & 58.38 & 71.91 & 90.53 & 61.07 & 69.85 & 84.52 \\
        \quad Qwen3-4B & Pre-Trained & 30.04 & 38.99 & 57.45 & 54.87 & 70.17 & 88.31 & 55.79 & 66.64 & 86.84 \\
        \quad Qwen3-30B-A3B & Pre-Trained & 43.39 & 53.42 & 72.32 & 61.40 & 75.90 & 93.59 & 64.22 & 73.64 & 90.31 \\
        \midrule
        \rowcolor{gray!20}\multicolumn{11}{l}{\textit{API}} \\
        \quad GPT-5-mini & API & 52.17 & \uwave{67.81} & \uwave{89.33} & 61.28 & 79.82 & \uwave{98.16} & \uwave{70.10} & \uwave{80.40} & 97.33 \\
        \quad DeepSeek-V3.2 & API & 52.15 & 66.91 & 83.34 & 62.77 & \uwave{79.85} & 96.97 & 68.12 & 78.04 & \uwave{97.42} \\
        \quad Gemini-2.5-Flash & API & \uwave{52.73} & 63.68 & 81.98 & \uwave{64.56} & 79.47 & 95.88 & 70.04 & 79.30 & 96.96 \\
        \midrule
        \rowcolor{gray!20}\multicolumn{11}{l}{\textit{SFT}} \\
        \quad Qwen2.5-0.5B & Trained & 1.31 & 6.15 & 6.07 & 13.32 & 22.26 & 19.45 & 21.77 & 26.78 & 16.50 \\
        \quad Qwen2.5-1.5B & Trained & 8.48 & 16.59 & 15.05 & 35.91 & 48.28 & 44.26 & 35.63 & 43.70 & 38.41 \\
        \quad Qwen2.5-3B & Trained & 20.15 & 30.11 & 42.36 & 49.46 & 64.05 & 80.04 & 53.59 & 62.57 & 71.23 \\
        \quad Qwen2.5-7B & Trained & 34.55 & 46.19 & 60.29 & 58.59 & 73.20 & 88.45 & 66.06 & 74.76 & 85.71 \\
        \midrule
        \rowcolor{gray!20}\multicolumn{11}{l}{\textit{Ours (RL)}} \\
        \quad CRAFT$_{\text{0.5B}}$ & Trained & 0.00 & 0.00 & 41.18 & 0.06 & 0.22 & 41.24 & 0.00 & 0.00 & 50.20 \\
        \quad CRAFT$_{\text{1.5B}}$ & Trained & 34.77 & 44.94 & 29.08 & 52.06 & 67.78 & 52.36 & 52.89 & 60.20 & 51.26 \\
        \quad CRAFT$_{\text{3B}}$ & Trained & 31.88 & 41.09 & 0.75 & 53.13 & 66.20 & 7.29 & 54.87 & 61.91 & 27.38 \\
        \quad CRAFT$_{\text{7B}}$ & Trained & \underline{51.86} & \underline{62.37} & \underline{84.60} & \underline{63.81} & \underline{78.55} & \underline{96.23} & \underline{76.62} & \underline{83.23} & \underline{95.97} \\
        \bottomrule
    \end{tabular*}
    \caption{Main results for \textsc{CRAFT}$_{\text{v3}}$ across three multi-hop QA benchmarks. This variant removes explicit document citation while retaining planning, reasoning, and answer extraction. EM/F1 and binary overall-consistency Faithfulness are reported in percentage. \uwave{Wavy underline}: best among API models; \underline{straight underline}: best among all other models.}
    \label{tab:main_results_v3}
\end{table*}

\begin{table*}[t]
    \centering
    \scriptsize
    \setlength{\tabcolsep}{3pt}
    \begin{tabular*}{\textwidth}{@{\extracolsep{\fill}}lcccccccccc@{}}
        \toprule
        \textbf{Method} & \textbf{Type} & \multicolumn{3}{c}{\textbf{MuSiQue}} & \multicolumn{3}{c}{\textbf{HotpotQA}} & \multicolumn{3}{c}{\textbf{2WikiMHQA}} \\
         &  & EM & F1 & Faith. & EM & F1 & Faith. & EM & F1 & Faith. \\
        \midrule
        \rowcolor{gray!20}\multicolumn{11}{l}{\textit{Open-source}} \\
        \quad Qwen2.5-0.5B & Pre-Trained & 0.39 & 2.62 & 5.00 & 2.74 & 7.51 & 13.76 & 4.85 & 10.09 & 8.02 \\
        \quad Qwen2.5-1.5B & Pre-Trained & 4.24 & 9.67 & 16.87 & 27.36 & 36.88 & 44.50 & 21.26 & 28.33 & 31.48 \\
        \quad Qwen2.5-3B & Pre-Trained & 21.91 & 32.32 & 41.55 & 48.93 & 61.63 & 76.24 & 41.46 & 50.78 & 60.12 \\
        \quad Qwen2.5-7B & Pre-Trained & 38.41 & 49.14 & 76.53 & 60.12 & 74.40 & 93.72 & 60.80 & 70.37 & 86.65 \\
        \quad Qwen3-4B & Pre-Trained & 38.42 & 47.69 & 77.53 & 60.66 & 76.05 & 95.49 & 65.32 & 74.45 & 93.80 \\
        \quad Qwen3-30B-A3B & Pre-Trained & 46.85 & 56.32 & 85.28 & \underline{63.40} & \underline{78.20} & \underline{96.88} & 70.69 & 78.98 & \underline{96.12} \\
        \midrule
        \rowcolor{gray!20}\multicolumn{11}{l}{\textit{API}} \\
        \quad GPT-5-mini & API & 52.38 & 68.00 & \uwave{92.87} & 60.84 & \uwave{79.84} & \uwave{98.12} & 69.83 & \uwave{81.01} & 97.18 \\
        \quad DeepSeek-V3.2 & API & \uwave{53.10} & \uwave{68.16} & 87.45 & 61.83 & 78.47 & 98.10 & \uwave{71.24} & 79.16 & \uwave{98.79} \\
        \quad Gemini-2.5-Flash & API & 51.38 & 62.90 & 89.62 & \uwave{63.36} & 79.80 & 97.10 & 70.70 & 79.21 & 97.86 \\
        \midrule
        \rowcolor{gray!20}\multicolumn{11}{l}{\textit{SFT}} \\
        \quad Qwen2.5-0.5B & Trained & 0.62 & 4.76 & 14.52 & 11.36 & 20.97 & 37.96 & 22.37 & 28.59 & 26.71 \\
        \quad Qwen2.5-1.5B & Trained & 8.22 & 13.53 & 16.54 & 34.76 & 45.49 & 53.73 & 36.59 & 44.12 & 43.17 \\
        \quad Qwen2.5-3B & Trained & 22.38 & 31.82 & 48.88 & 50.25 & 63.74 & 84.38 & 54.69 & 63.34 & 73.35 \\
        \quad Qwen2.5-7B & Trained & 30.11 & 41.53 & 65.80 & 57.54 & 71.84 & 89.40 & 63.92 & 72.97 & 82.37 \\
        \midrule
        \rowcolor{gray!20}\multicolumn{11}{l}{\textit{Ours (RL)}} \\
        \quad CRAFT$_{\text{0.5B}}$ & Trained & 0.00 & 2.15 & \underline{85.43} & 0.00 & 3.97 & 91.82 & 0.00 & 4.21 & 85.70 \\
        \quad CRAFT$_{\text{1.5B}}$ & Trained & 34.15 & 44.93 & 55.74 & 47.42 & 62.14 & 57.90 & 48.92 & 56.75 & 55.93 \\
        \quad CRAFT$_{\text{3B}}$ & Trained & 34.63 & 43.49 & 66.41 & 56.72 & 71.40 & 87.12 & 61.57 & 69.78 & 76.41 \\
        \quad CRAFT$_{\text{7B}}$ & Trained & \underline{48.02} & \underline{59.94} & 83.17 & 62.64 & 77.84 & 94.96 & \underline{75.60} & \underline{82.04} & 92.90 \\
        \bottomrule
    \end{tabular*}
    \caption{Main results for \textsc{CRAFT}$_{\text{v4}}$ across three multi-hop QA benchmarks. This variant retains only reasoning and answer extraction, providing a compact structured trace. EM/F1 and binary overall-consistency Faithfulness are reported in percentage. \uwave{Wavy underline}: best among API models; \underline{straight underline}: best among all other models.}
    \label{tab:main_results_v4}
\end{table*}

\begin{table*}[ht!]
    \centering
    \scriptsize
    \setlength{\tabcolsep}{4pt}
    \begin{tabular*}{\textwidth}{@{\extracolsep{\fill}}lccccccc@{}}
        \toprule
        \textbf{Method} & \textbf{Type} & \multicolumn{2}{c}{\textbf{MuSiQue}} & \multicolumn{2}{c}{\textbf{HotpotQA}} & \multicolumn{2}{c}{\textbf{2WikiMHQA}} \\
         &  & EM & F1 & EM & F1 & EM & F1 \\
        \midrule
        \rowcolor{gray!20}\multicolumn{8}{l}{\textit{Open-source}} \\
        \quad Qwen2.5-0.5B & Pre-Trained & 0.40 & 3.23 & 5.50 & 12.02 & 6.32 & 12.55 \\
        \quad Qwen2.5-1.5B & Pre-Trained & 9.67 & 17.63 & 35.75 & 47.66 & 26.23 & 33.32 \\
        \quad Qwen2.5-3B & Pre-Trained & 23.38 & 33.23 & 49.73 & 62.52 & 41.25 & 50.25 \\
        \quad Qwen2.5-7B & Pre-Trained & 31.43 & 41.91 & 53.38 & 67.97 & 51.01 & 59.21 \\
        \quad Qwen3-4B & Pre-Trained & 39.64 & 49.19 & 60.19 & 75.30 & 57.11 & 67.29 \\
        \quad Qwen3-30B-A3B & Pre-Trained & 48.78 & 59.31 & \underline{62.79} & \underline{78.33} & \underline{64.37} & \underline{73.61} \\
        \midrule
        \rowcolor{gray!20}\multicolumn{8}{l}{\textit{API}} \\
        \quad GPT-5-mini & API & 51.22 & 65.02 & 63.59 & 80.55 & 64.62 & 76.69 \\
        \quad DeepSeek-V3.2 & API & \uwave{58.51} & \uwave{70.72} & 64.35 & 80.37 & 66.36 & 76.47 \\
        \quad Gemini-2.5-Flash & API & 55.48 & 67.22 & \uwave{66.11} & \uwave{81.03} & \uwave{71.54} & \uwave{80.78} \\
        \midrule
        \rowcolor{gray!20}\multicolumn{8}{l}{\textit{SFT}} \\
        \quad Qwen2.5-0.5B & Trained & 0.13 & 0.91 & 0.78 & 2.99 & 2.87 & 4.84 \\
        \quad Qwen2.5-1.5B & Trained & 10.30 & 17.38 & 33.16 & 46.02 & 33.71 & 41.19 \\
        \quad Qwen2.5-3B & Trained & 18.53 & 27.16 & 44.65 & 59.07 & 50.56 & 58.17 \\
        \quad Qwen2.5-7B & Trained & 25.60 & 36.27 & 54.30 & 68.26 & 55.47 & 63.41 \\
        \midrule
        \rowcolor{gray!20}\multicolumn{8}{l}{\textit{Ours (RL)}} \\
        \quad CRAFT$_{\text{0.5B}}$ & Trained & 23.87 & 33.30 & 36.70 & 50.23 & 36.65 & 43.71 \\
        \quad CRAFT$_{\text{1.5B}}$ & Trained & 34.75 & 44.25 & 50.23 & 66.07 & 46.34 & 53.13 \\
        \quad CRAFT$_{\text{3B}}$ & Trained & 43.59 & 53.71 & 59.65 & 73.77 & 53.64 & 62.13 \\
        \quad CRAFT$_{\text{7B}}$ & Trained & \underline{49.95} & \underline{60.51} & 59.80 & 74.79 & 60.34 & 67.72 \\
        \bottomrule
    \end{tabular*}
    \caption{Main results for \textsc{CRAFT}$_{\text{v5}}$ across three multi-hop QA benchmarks. This answer-only variant removes structured reasoning fields, so Faithfulness is not applicable and only EM/F1 are reported. \uwave{Wavy underline}: best among API models; \underline{straight underline}: best among all other models.}
    \label{tab:main_results_v5}
\end{table*}

\begin{table*}[t]
    \centering
    \scriptsize
    \setlength{\tabcolsep}{1.8pt}
    \begin{tabular*}{\textwidth}{@{\extracolsep{\fill}}llcccccccccc@{}}
        \toprule
        \multirow{2}{*}{\textbf{Model}} & \multirow{2}{*}{\textbf{Dataset}} &
\multicolumn{2}{c}{\textbf{CRAFT$_{\text{v1}}$}} & \multicolumn{2}{c}{\textbf{CRAFT$_{\text{v2}}$}} &
\multicolumn{2}{c}{\textbf{CRAFT$_{\text{v3}}$}} & \multicolumn{2}{c}{\textbf{CRAFT$_{\text{v4}}$}} &
\multicolumn{2}{c}{\textbf{CRAFT$_{\text{v5}}$}} \\
        \cmidrule(lr){3-4} \cmidrule(lr){5-6} \cmidrule(lr){7-8} \cmidrule(lr){9-10} \cmidrule(lr){11-12}
        & & Base & GRPO & Base & GRPO & Base & GRPO & Base & GRPO & Base & GRPO \\
        \midrule
        \rowcolor{gray!20}\multicolumn{12}{l}{\textit{Qwen2.5-0.5B}} \\
        & Avg. & 0.00 & \textbf{24.97} & 1.12 & \textbf{95.77} & 0.03 & \textbf{0.00} & 1.03 & \textbf{99.50} & 9.02 & \textbf{100.00} \\
        & HotpotQA & 0.00 & 25.00 & 0.10 & 95.20 & 0.10 & 0.00 & 0.80 & 99.10 & 5.65 & 100.00 \\
        & 2WikiMHQA & 0.00 & 35.70 & 3.25 & 94.90 & 0.00 & 0.00 & 2.30 & 99.80 & 16.50 & 100.00 \\
        & MuSiQue & 0.00 & 14.20 & 0.00 & 97.20 & 0.00 & 0.00 & 0.00 & 99.60 & 4.90 & 100.00 \\
        \midrule
        \rowcolor{gray!20}\multicolumn{12}{l}{\textit{Qwen2.5-1.5B}} \\
        & Avg. & 0.05 & \textbf{91.90} & 2.12 & \textbf{87.53} & 0.22 & \textbf{96.30} & 1.20 & \textbf{94.17} & 0.75 & \textbf{100.00} \\
        & HotpotQA & 0.00 & 90.80 & 2.95 & 95.90 & 0.05 & 96.40 & 1.00 & 93.20 & 0.70 & 100.00 \\
        & 2WikiMHQA & 0.00 & 92.80 & 0.80 & 92.90 & 0.00 & 98.10 & 1.40 & 94.50 & 1.35 & 100.00 \\
        & MuSiQue & 0.15 & 92.10 & 2.60 & 73.80 & 0.60 & 94.40 & 1.20 & 94.80 & 0.20 & 100.00 \\
        \midrule
        \rowcolor{gray!20}\multicolumn{12}{l}{\textit{Qwen2.5-3B}} \\
        & Avg. & 13.27 & \textbf{96.27} & 50.28 & \textbf{99.30} & 6.43 & \textbf{15.60} & 15.57 & \textbf{98.83} & 41.62 & \textbf{100.00} \\
        & HotpotQA & 14.80 & 95.00 & 55.25 & 99.40 & 4.30 & 9.60 & 27.60 & 98.90 & 32.80 & 100.00 \\
        & 2WikiMHQA & 7.70 & 98.90 & 33.80 & 99.80 & 6.30 & 35.70 & 8.20 & 99.70 & 23.95 & 100.00 \\
        & MuSiQue & 17.30 & 94.90 & 61.80 & 98.70 & 8.70 & 1.50 & 10.90 & 97.90 & 68.10 & 100.00 \\
        \midrule
        \rowcolor{gray!20}\multicolumn{12}{l}{\textit{Qwen2.5-7B}} \\
        & Avg. & 94.13 & \textbf{98.98} & 99.20 & \textbf{99.62} & 94.40 & \textbf{98.50} & 61.65 & \textbf{98.55} & 15.35 & \textbf{100.00} \\
        & HotpotQA & 96.10 & 99.45 & 99.45 & 99.85 & 95.65 & 99.40 & 49.50 & 99.45 & 11.15 & 100.00 \\
        & 2WikiMHQA & 95.40 & 99.80 & 99.30 & 99.60 & 95.95 & 99.45 & 72.90 & 98.65 & 9.55 & 100.00 \\
        & MuSiQue & 90.90 & 97.70 & 98.85 & 99.40 & 91.60 & 96.65 & 62.55 & 97.55 & 25.35 & 100.00 \\
        \bottomrule
    \end{tabular*}
    \caption{Measured format compliance (\%) across CRAFT variants, model scales, and datasets. ``Avg.'' is the arithmetic mean over MuSiQue, HotpotQA, and 2WikiMHQA. GRPO improves schema compliance across most settings; \textsc{CRAFT}$_{\text{v3}}$ remains the most challenging structured template at 0.5B and 3B.}
    \label{tab:format_accuracy}
\end{table*}

\subsection{Full Results for Additional Variants}
\label{sec:app_full_results}

Tables~\ref{tab:main_results_v2}--\ref{tab:main_results_v5} present the complete evaluation results for \textsc{CRAFT}$_{\text{v2}}$--\textsc{CRAFT}$_{\text{v5}}$ across all model scales and baselines, complementing the \textsc{CRAFT}$_{\text{v1}}$ results in the main paper (Table~\ref{tab:main_results_v1}).

\noindent\textbf{CRAFT$_{\text{v2}}$ Results.}
CRAFT$_{\text{v2}}$ retains the same structure as CRAFT$_{\text{v1}}$ but removes \texttt{<plan>}. CRAFT$_{\text{7B}}$ achieves 51.31\% EM on MuSiQue (+17.07 points vs.\ Base), 65.28\% on HotpotQA (+10.72), and 75.26\% on 2WikiMHQA (+19.66). It surpasses all three API models in EM on HotpotQA and 2WikiMHQA while reaching 82.49\%--96.38\% Faithfulness.

\noindent\textbf{CRAFT$_{\text{v3}}$ Results.}
CRAFT$_{\text{v3}}$ removes the explicit citation field (\texttt{<gold\_docs>}) and $R_{\mathrm{gold}}$, while retaining judge checks over the plan, reasoning, and answer. CRAFT$_{\text{7B}}$ achieves 51.86\% EM on MuSiQue, 63.81\% on HotpotQA, and 76.62\% on 2WikiMHQA. The model reaches 84.60\%--96.23\% Faithfulness, showing that plan--reason--answer consistency retains substantial auditable structure even without explicit document identifiers.

\noindent\textbf{CRAFT$_{\text{v4}}$ Results.}
CRAFT$_{\text{v4}}$ retains only \texttt{<reason>}+\texttt{<answer>}. It reaches 83.17\%/94.96\%/92.90\% Faithfulness and 48.02\%/62.64\%/75.60\% EM on MuSiQue/HotpotQA/2WikiMHQA. Across the trained 7B variants, v3 is highest in Faithfulness on MuSiQue, v2 on HotpotQA, and v1 on 2WikiMHQA, showing that the compact v4 trace remains competitive but is not uniformly best.

\noindent\textbf{CRAFT$_{\text{v5}}$ Results.}
CRAFT$_{\text{v5}}$ uses direct answer generation without structured reasoning traces ($R_{\mathrm{faith}}$ and $R_{\mathrm{gold}}$ not applicable). At 7B, it achieves 49.95\% EM on MuSiQue, 59.80\% on HotpotQA, and 60.34\% on 2WikiMHQA. At 0.5B, v5 reaches 23.87\%/36.70\%/36.65\% EM, far above the structured variants, indicating that direct answer optimization is substantially easier for the smallest model.

\begin{table}[t]
\centering
\scriptsize
\setlength{\tabcolsep}{1.5pt}
\begin{tabular}{|p{0.10\columnwidth}|p{0.40\columnwidth}|p{0.40\columnwidth}|}
\hline
\textbf{Version} & \textbf{Base Model Errors (7B)} & \textbf{GRPO Model Errors (7B)} \\
\hline
\rowcolor{gray!10}
\textbf{v1} & \textcolor{red}{\textbf{Truncated reason}}: \newline response ends at \texttt{[Answer} \newline \textcolor{red}{$\leftarrow$ missing \texttt{</reason>} and answer block} & \textcolor{red}{\textbf{Runaway reason}}: \newline repeated text reaches the length limit \newline \textcolor{red}{$\leftarrow$ missing closing fields} \\
\hline
\rowcolor{gray!10}
\textbf{v2} & \textcolor{red}{\textbf{Tag mismatch}}: \newline \texttt{<answer>[Rabbi Dovber Schneuri]} \newline closed with \texttt{[/answer]} instead of \texttt{</answer>} & \textcolor{green!50!black}{\textbf{Near-perfect compliance \checkmark}} \\
\hline
\rowcolor{gray!10}
\textbf{v3} & \textcolor{red}{\textbf{Truncated reason}}: \newline response ends before \texttt{</reason>} \newline \textcolor{red}{$\leftarrow$ answer block absent} & \textcolor{red}{\textbf{Runaway reason}}: \newline repeated text reaches the length limit \newline \textcolor{red}{$\leftarrow$ answer block absent} \\
\hline
\rowcolor{gray!10}
\textbf{v4} & \textcolor{red}{\textbf{Missing closing tag}}: \newline \texttt{<answer>1986} \newline \textcolor{red}{$\leftarrow$ missing \texttt{</answer>}} & \textcolor{red}{\textbf{Extra text}}: \newline generated text continues after \texttt{</answer>} \\
\hline
\rowcolor{gray!10}
\textbf{v5} & \textcolor{red}{\textbf{Wrong format}}: \newline bare text without XML tags \newline e.g., \texttt{[1493]} & \textcolor{green!50!black}{\textbf{No format failures \checkmark}} \\
\hline
\end{tabular}
\caption{Representative 7B format behavior across template versions. GRPO achieves 98.50\%--100.00\% average compliance; the displayed GRPO failures are residual cases.}
\label{tab:format_errors}
\end{table}
\subsection{Format Compliance Analysis}
\label{sec:format_analysis}

Expanding on the format compliance summary in \S\ref{sec:training_dynamics}, Table~\ref{tab:format_accuracy} presents measured compliance on 2,000 randomly sampled examples from each dataset for every CRAFT variant and model scale.

\noindent\textbf{GRPO substantially improves format compliance across most settings.} Base models struggle with structured output, particularly at smaller scales: 0.5B and 1.5B base models achieve near-zero format accuracy on complex templates, while 3B Base averages range from 6.43\% to 50.28\% on CRAFT$_{\text{v1}}$--CRAFT$_{\text{v4}}$. For 0.5B, the GRPO changes on these four templates are +24.97, +94.65, $-$0.03, and +98.47 percentage points; the v3 exception remains at 0.00\%. The corresponding improvement ranges are 85.42--96.08 points for 1.5B, 9.17--83.27 points for 3B, and 0.42--36.90 points for 7B. The 7B GRPO models reach 98.50\%--99.62\% compliance across CRAFT$_{\text{v1}}$--CRAFT$_{\text{v4}}$.

\noindent\textbf{Format compliance alone does not imply answer quality.} CRAFT$_{\text{0.5B}}$ reaches 99.50\% format compliance on v4 but remains at 0.00\% EM on all three datasets, so it learns the schema without learning correct answers. In contrast, answer-only v5 reaches 100.00\% compliance together with 23.87\%--36.70\% EM. CRAFT$_{\text{0.5B}}$-v3 has 0.00\% format compliance, and CRAFT$_{\text{3B}}$-v3 reaches only 15.60\%; the latter agrees with the low v3 Faithfulness in Table~\ref{tab:main_results_v3}. At 7B, all five variants achieve 96.65\%--100.00\% dataset-level format compliance, indicating that the remaining performance differences are not explained by schema adherence alone.

\subsection{Format Error Analysis}
\label{sec:app_format_error}

At 7B, GRPO reaches 98.50\%--100.00\% average format compliance across all five variants (Table~\ref{tab:format_accuracy}). Table~\ref{tab:format_errors} therefore focuses on the small residual error set, including occasional missing closing tags, response truncation, and extra text, while \textsc{CRAFT}$_{\text{v2}}$ and \textsc{CRAFT}$_{\text{v5}}$ are near-perfect or perfect in the measured cohort.

\subsection{Capacity-Dependent Template Preference}
\label{sec:app_05b}

The 0.5B results reveal a clear capacity-dependent template preference. SFT obtains its strongest structured-template performance with \textsc{CRAFT}$_{\text{v3}}$ and \textsc{CRAFT}$_{\text{v4}}$, whereas GRPO favors the answer-only \textsc{CRAFT}$_{\text{v5}}$, which reaches 23.87/36.70/36.65\% EM on MuSiQue/HotpotQA/2WikiMHQA and substantially outperforms the structured variants at this scale.

This pattern yields a practical operating point: \textbf{\textsc{CRAFT}$_{\text{v5}}$ is the most effective 0.5B configuration for answer accuracy}. Removing intermediate generation requirements concentrates the smallest model's capacity on answer optimization; at larger scales, structured variants recover strong answer performance while providing richer auditability.

\section{Case Studies}
\label{sec:case_studies_appendix}
These examples illustrate two complementary failure modes that CRAFT addresses: right-answer-wrong-reason and wrong-answer-wrong-reason. In both cases, \textsc{CRAFT}$_{\text{7B}}$ restores faithful evidence use and receives an all-checks-pass audit.

\subsection{\texorpdfstring{Case: Base (EM=1, $F_{\mathrm{eval}}$=0) $\rightarrow$ CRAFT (EM=1, $F_{\mathrm{eval}}$=1)}{Case: Base (EM=1, eval=0) to CRAFT (EM=1, eval=1)}}
\label{sec:case_study}

Table~\ref{tab:case_study} presents a MuSiQue case where the base model gives the correct answer through unfaithful reasoning. It answers ``\$59,039'' but confuses Aaron Spelling with Lorraine Crosby, concludes that the creator is English, and then uses US income data from outside its declared evidence set. The audit catches this distractor-induced entity confusion and citation-boundary violation, while \textsc{CRAFT}$_{\text{7B}}$ traces Spelling$\rightarrow$American$\rightarrow$US income and passes all four checks.

\begin{table}[t!]
\centering
\scriptsize
\setlength{\tabcolsep}{1.5pt}
\begin{tabular}{|p{0.13\columnwidth}|p{0.4\columnwidth}|p{0.4\columnwidth}|}
\hline
\textbf{Aspect} & \textbf{Base (EM=1, Faith=0)} & \textbf{CRAFT$_{\text{7B}}$ (EM=1, Faith=1)} \\
\hline
\rowcolor{plancolor!40}
\textbf{Plan} & Subq1: Who is the creator? \newline Subq2: What is the nationality? \newline Subq3: What is the avg income? & Subq1: Who is the creator? \newline Subq2: What is the creator's nationality? \newline Subq3: What is the avg income for that nationality? \\
\hline
\rowcolor{golddoccolor!40}
\textbf{Gold Docs} & [7, 13] & [9, 7, 1] \\
\hline
\rowcolor{reasoncolor!40}
\textbf{Reason} & Identifies Aaron Spelling (doc [7]). \textcolor{red}{\textbf{Confuses him with Lorraine Crosby (doc [13]), concludes ``English'', then cites out-of-boundary doc [1] for US income.}} & Doc [9]: Spelling created MacGruder and Loud. Doc [7]: Spelling is American. Doc [1]: US average salary is \$59,039. \\
\hline
\rowcolor{answercolor!60}
\textbf{Answer} & \$59,039 \checkmark~\textit{(correct but unfaithful)} & \textcolor{green!50!black}{\textbf{\$59,039 \checkmark}} \\
\hline
\hline
\multicolumn{3}{|c|}{\cellcolor{gray!20}\textbf{Judge Evaluation}} \\
\hline
$c_\pi$ & \textcolor{red}{0} - Concludes ``English'' despite identifying Spelling & \textcolor{green!50!black}{1} - creator$\rightarrow$nationality$\rightarrow$income \\
\hline
$c_\mathcal{E}$ & \textcolor{red}{0} - Cites doc [1], outside gold\_docs [7, 13] & \textcolor{green!50!black}{1} - All cited docs in gold\_docs \\
\hline
$c_a$ & \textcolor{red}{0} - Derives ``English'' but uses US income & \textcolor{green!50!black}{1} - \$59,039 supported by reasoning \\
\hline
$c_g$ & \textcolor{red}{0} - ``Creator is English'' unsupported & \textcolor{green!50!black}{1} - All claims verified \\
\hline
$F_\mathrm{eval}$ & \textcolor{red}{\textbf{0}} & \textcolor{green!50!black}{\textbf{1}} \\
\hline
\end{tabular}
\caption{Case study: Base model produces the correct answer through unfaithful reasoning, while \textsc{CRAFT}$_{\text{7B}}$ achieves both correctness and faithfulness.}
\label{tab:case_study}
\end{table}
\begin{table}[ht!]
\centering
\scriptsize
\setlength{\tabcolsep}{1.5pt}
\begin{tabular}{|p{0.13\columnwidth}|p{0.4\columnwidth}|p{0.4\columnwidth}|}
\hline
\textbf{Aspect} & \textbf{Base Model (EM=0, $F_{\mathrm{eval}}$=0)} & \textbf{CRAFT$_{\text{7B}}$ (EM=1, $F_{\mathrm{eval}}$=1)} \\
\hline
\rowcolor{plancolor!40}
\textbf{Plan} & Subq1: Who released Holiday Harmony? \newline Subq2: What type of radio stations played that group's songs? & Subq1: Which group released Holiday Harmony? \newline Subq2: What type of radio stations played that group's songs? \\
\hline
\rowcolor{golddoccolor!40}
\textbf{Gold Docs} & [6, 7] & [6, 7] \\
\hline
\rowcolor{reasoncolor!40}
\textbf{Reason} & Docs [6, 7] identify America as the group. \textcolor{red}{\textbf{The trace then cites undeclared doc [8] about another group's song to infer Adult Contemporary airplay.}} & Doc [6]: America released Holiday Harmony. Doc [7]: America's songs received airplay on pop/soft rock stations. \\
\hline
\rowcolor{answercolor!60}
\textbf{Answer} & \textcolor{red}{\textbf{Adult Contemporary}} & \textcolor{green!50!black}{\textbf{pop/soft rock stations \checkmark}} \\
\hline
\hline
\multicolumn{3}{|c|}{\cellcolor{gray!20}\textbf{Judge Evaluation}} \\
\hline
$c_\pi$ & \textcolor{green!50!black}{1} - Plan steps are present & \textcolor{green!50!black}{1} - Plan followed \\
\hline
$c_\mathcal{E}$ & \textcolor{red}{0} - Cites doc [8], outside gold\_docs [6, 7] & \textcolor{green!50!black}{1} - All citations stay within gold\_docs \\
\hline
$c_a$ & \textcolor{green!50!black}{1} - Answer follows the stated chain & \textcolor{green!50!black}{1} - Answer follows the stated chain \\
\hline
$c_g$ & \textcolor{red}{0} - Doc [8] concerns a different group & \textcolor{green!50!black}{1} - All claims verified \\
\hline
$R_\mathrm{faith}^{\mathrm{train}}$ & \textcolor{red}{\textbf{0.50}} & \textcolor{green!50!black}{\textbf{1.00}} \\
\hline
$F_\mathrm{eval}$ & \textcolor{red}{\textbf{0}} & \textcolor{green!50!black}{\textbf{1}} \\
\hline
\end{tabular}
\caption{Case study: the Base model combines an out-of-boundary citation with unsupported evidence and returns an incorrect answer, while CRAFT recovers both correctness and an all-checks-pass audit.}
\label{tab:case1_appendix}
\end{table}
\subsection{\texorpdfstring{Case: Base (EM=0, $F_{\mathrm{eval}}$=0) $\rightarrow$ CRAFT (EM=1, $F_{\mathrm{eval}}$=1)}{Case: Base (EM=0, eval=0) to CRAFT (EM=1, eval=1)}}
\label{sec:case_study_hotpot}

\noindent\textbf{Question:} What type of radio stations played songs by the group who released the album Holiday Harmony?

\noindent\textbf{Gold Answer:} pop/soft rock stations

Table~\ref{tab:case1_appendix} presents an additional HotpotQA case where both accuracy and faithfulness are corrected. The base model combines an out-of-boundary citation with unsupported evidence and returns the wrong answer, while \textsc{CRAFT}$_{\text{7B}}$ stays within the declared evidence and recovers the correct answer. This case also illustrates the distinction between component-average training reward ($0.50$ from two of four passing checks) and the binary evaluation verdict ($0$).

Together, the two cases show that CRAFT's dual reward design improves both answer correctness and reasoning faithfulness, including cases where answer-only metrics would miss the underlying failure.

\end{document}